\documentclass[a4paper,11pt,final]{article}

\usepackage{booktabs} % For formal tables
\usepackage{amsmath}
\usepackage{amssymb}
\usepackage[ruled,vlined,linesnumbered]{algorithm2e}

\newcommand{\ignore}[1]{}
\usepackage[noend]{algpseudocode}
\usepackage[T1]{fontenc}
\usepackage{tabularx}
\usepackage{multirow}
\usepackage{array}
\usepackage{pbox}
\usepackage{balance}
\usepackage{color,soul}
\usepackage{nameref}
\pdfmapfile{+lmodern.map}
\usepackage{microtype}
\usepackage{pgfplots}
\usepackage{graphicx}
\usepackage{subcaption}
\usepackage{tikz}
\usepackage{siunitx}
\usepackage{url}
\usepackage{bm}

\usepackage[normalem]{ulem}
\begin{document}

\title{\ \\ \LARGE\bf Evolutionary Algorithms for Generating Graphs Matching Desired Laplacian Spectra}

\author{ \Large Hendrik Richter$^{1}$ and Frank Neumann$^{2}$ \\ \\
	$^{1}$ HTWK Leipzig
University of Applied Sciences, \\ Faculty of  Engineering, D–04251 Leipzig, Germany\\
	$^{2}$ Optimisation and Logistics, Adelaide University,\\ Adelaide, SA 5005, Australia }

	\maketitle

    \begin{abstract}

Graphs with diverse structural characteristics play a central role in modelling and optimization tasks. The ability to generate different types of graphs that exhibit shared properties is likewise essential for algorithm selection and configuration. However, constructing graphs that preserve high-level properties across a broad range of graph classes remains a challenging problem.
We present a novel evolutionary approach to evolve graphs based on the Laplacian graph spectra descriptor. 
This descriptor can be used as part of a fitness function to evaluate graphs according to their desired high-level properties.
Our evolutionary algorithm evolves graphs towards this descriptor in order to obtain graphs having properties that are consistent with it but are different from each other in terms of non-spectral graph metrics, such as path length, clustering coefficient and betweenness centrality. 
Our experimental results show that our approach is successful for different classes of graphs and a wide range of Laplacian graph spectra. % descriptors.
    \end{abstract}

\section{Introduction}

%Evolutionary algorithms are general purpose algorithms that have found applications in a wide range of domains and are especially well suited to deal with complex objective functions. 
Representing concepts and relationships through graphs is a central paradigm in computer science and mathematics. Thus, generating graphs with distinct properties facilitates  analyzing  graph algorithms and yields important insights into  structural features.

In network science and machine learning, for instance, algorithms for generating graphs with desired structures are  important in applications such as evaluating computational performance and correctness of network algorithms or protocols, or data augmentation, benchmarking, and model interpretation of graph neural networks. Examples are random-based generating algorithms of Erdös-Renyi~\cite{er59}, Barabasi-Albert~\cite{alb02} and Watts-Strogatz~\cite{watts98}, but also deep graph generators~\cite{fa21,guo22,boni20}. Typically, the parameters of these algorithms allow mainly to adjust local network characteristics such as degree, path length and centrality. 

However, relevant network features can also be described by global network properties. Recently, it has been shown that the spectrum of the (normalized) Laplacian is particularly promising for such a global characterization, as it naturally encodes high-level connectivity information~\cite{shi19,ban09,rich21,rich23,wills20}, for instance, 
the strength of connectivity and expansion, the
presence of clusters, the distance to bipartiteness,
 and the nature of
degree-normalized structural patterns.   
Thus, the paper also encourages considering a shift from graph generation methods that replicate local properties to those that capture global connectivity structures.

\subsection{Related Work} \label{sec:related}
Different evolutionary approaches have been developed for generating graphs.
For example, genetic programming~\cite{atkin18,sotto20} has been widely employed to produce graphs for given application domains, for instance  digital circuit synthesis benchmarks. 
Other examples are in the context of evolving artificial neural networks  where neuroevolution of augmenting topologies techniques have been applied successfully~\cite{papa21}.

Other approaches for generating graphs by evolutionary computation focus on graph properties as objectives. Examples are
multi-objective genetic algorithms for producing graphs that simultaneously satisfy graph-topological properties such as degree distribution, self-loop abundance and connectance~\cite{leith24}, or  
 searching for graphs with specific graph-theoretical features, for instance, finding influential nodes~\cite{liu19}, or  generating dense families of circulant graphs~\cite{mon02},  or reconstructing graphs with a target  betweenness  centrality~\cite{loz21}, or identifying specific tree structures~\cite{med23}. There are also works which are
using templates to evolve graphs towards a given average vertex degree, degree distribution and  clustering coefficient~\cite{bach13,ver17}.

Although receiving attention in machine learning~\cite{cir22,luo23}, the use of Laplacian graph spectra as an optimization objective for graph generation has, to the best of our knowledge, not yet been addressed in evolutionary computation. The nearest precedent is incorporating spectral metrics of the adjacency matrix, namely graph energy (defined as the sum of the eigenvalues of the adjacency matrix), into evolutionary search~\cite{jim21}. 
A second related approach  employs  a Metropolis algorithm for reconstructing graphs from their Laplacian spectra~\cite{ibs02}. 

\subsection{Our Contribution} 
We provide a novel approach on the use of evolutionary algorithms for generating graphs that match given properties based on a Laplacian graph spectra descriptor. 
As a novelty, we use the Laplacian spectrum of the graph as the main component of the objective function and thus as a guiding principle in the evolutionary search. There are two objectives in our study. The first is to design evolutionary algorithms that successfully generate graphs matching Laplacian target spectra. The second is to produce graph diversity with respect to (non-spectral) graph metrics, for instance, (average) path length, (global) clustering coefficient, and (average) betweenness centrality. 

Our graph generation procedure enforces the graph size  to be fixed throughout the evolutionary process.
Thus, changes in the graph spectrum are induced by edge manipulations, which are carried out on a smaller (local) scale by mutation and on a larger (global) scale by crossover. Because the Laplacian spectrum is strongly influenced by degree patterns,  edge manipulations should be guided  for helping the evolving graph moving more effectively toward the target spectral characteristics.  
A main design principle in the algorithm's implementation is the use of spectral quantities for guiding and controlling mutation and crossover. By numerical experiments we can show that such an implementation is successful in meeting both objectives, generating graphs that match Laplacian target spectra and producing diversity in terms of non-spectral graph metrics.

The remainder of this paper is organized as follows. 
In Section \ref{sec:methods}, we recall the spectral graph description which is the key concept for our approach. In Section~\ref{sec:alg}, we present our algorithm for evolving graphs towards graph spectra.
Extensive experimental investigations are given in Section~\ref{sec:results}. 
We finish with some conclusions.

\section{Graphs and Graph Spectra} \label{sec:methods}

We consider an (undirected, unweighted) graph  $\mathcal{G}=(V,E)$ with a set $V$ of vertices $v_i \in V$, $i=1,2,\ldots,n$  and a set $E$ of edges $e_{ij}\in E$ connecting $v_i$ and $v_j$. For such a graph there is a matrix representation by the 
symmetric adjacency matrix $A=\{a_{ij}\}$, where $a_{ij}=a_{ji}=1$ indicates that the vertices $v_i$ and $v_j$ are connected by the edge $e_{ij}$. With the vertex degree $k_i=\sum_{j=1}^{n} a_{ij}$, we additionally find a degree matrix $D=\rm{diag}$ $(k_1,k_2,\ldots,k_n)$. 

For the  spectral description of a graph $\mathcal{G}$, we have $A$ and $D$, and get   the  normalized Laplacian $\Lambda_{\mathcal{G}}=I-D^{-1/2}AD^{-1/2}$.   Its eigenvalue spectrum is $\lambda(\mathcal{G})$ with $0=\lambda_1\leq \lambda_2 \leq \ldots \lambda_n\leq 2$. 
 We consider a smoothed spectral density obtained by convolving the eigenvalues  $\lambda_i$ with a Gaussian kernel centered at $x \in \mathbb{R}$ with standard deviation $\sigma$~\cite{gu16} 
\begin{equation}
    \varphi_{\mathcal{G}}(x)= \frac{1}{n}\sum_{i=1}^{n} \frac{1}{\sqrt{2 \pi \sigma^2}} \exp{\left(-\frac{(x-\lambda_i)^2}{2 \sigma^2} \right)} \label{eq:density}
\end{equation} and set $\sigma=1/n$.

\section{Evolving Graphs based on Graph Spectra}
\label{sec:alg}

We now outline our evolutionary algorithm for evolving graphs towards a given graph spectrum. 

\subsection{Representation and Initial Population}
The vertices and edges of the graph are formally represented by the elements of the symmetric adjacency matrix $A=\{a_{ij}\}$. However, the adjacency matrix is merely  a convenient means for specifying the graph, which could equivalently be done by an edge list. In particular, the matrix $A$ is not converted to a binary string as previous studies have shown that such a direct binary string representation is not very helpful for supporting the evolutionary search, particularly if the graphs are sparse~\cite{ash14}. 

The population has size $\ell$. It is constant for the run time of the evolutionary algorithm, which  uses  tournament selection with tournament size $s_t=2$, genetic operators (see Sec.~\ref{sec:genetic}), and elitism.    For initial populations we use different graph sets and study the effect of the choice on performance and graph diversity. On the one hand, we employ three groups of random graphs,  Erd\"os--Renyi (ER)~\cite{er59}, Barabasi-Albert (BA)~\cite{alb02} and Watts-Strogatz (WS)~\cite{watts98}. Moreover, we  
take subsets of pairwise nonisomorphic $k$-regular graphs, which can be generated algorithmically~\cite{mer99}. In particular, we use cubic ($k=3$), sextic ($k=6$), and nonic ($k=9$) regular graphs as well as $k=12$ and $k=16$. We test several graph size $n=\{24, 64, 128, 256, 512\}$. We require all initial graphs to be connected, and the genetic operators to be designed to ensure that all graphs in the population remain connected.

\subsection{Fitness Function and Measuring Performance}
\begin{figure}[t]

\begin{center}
\includegraphics[trim = 60mm 40mm 70mm 30mm,clip, width=6.2cm, height=4.2cm]{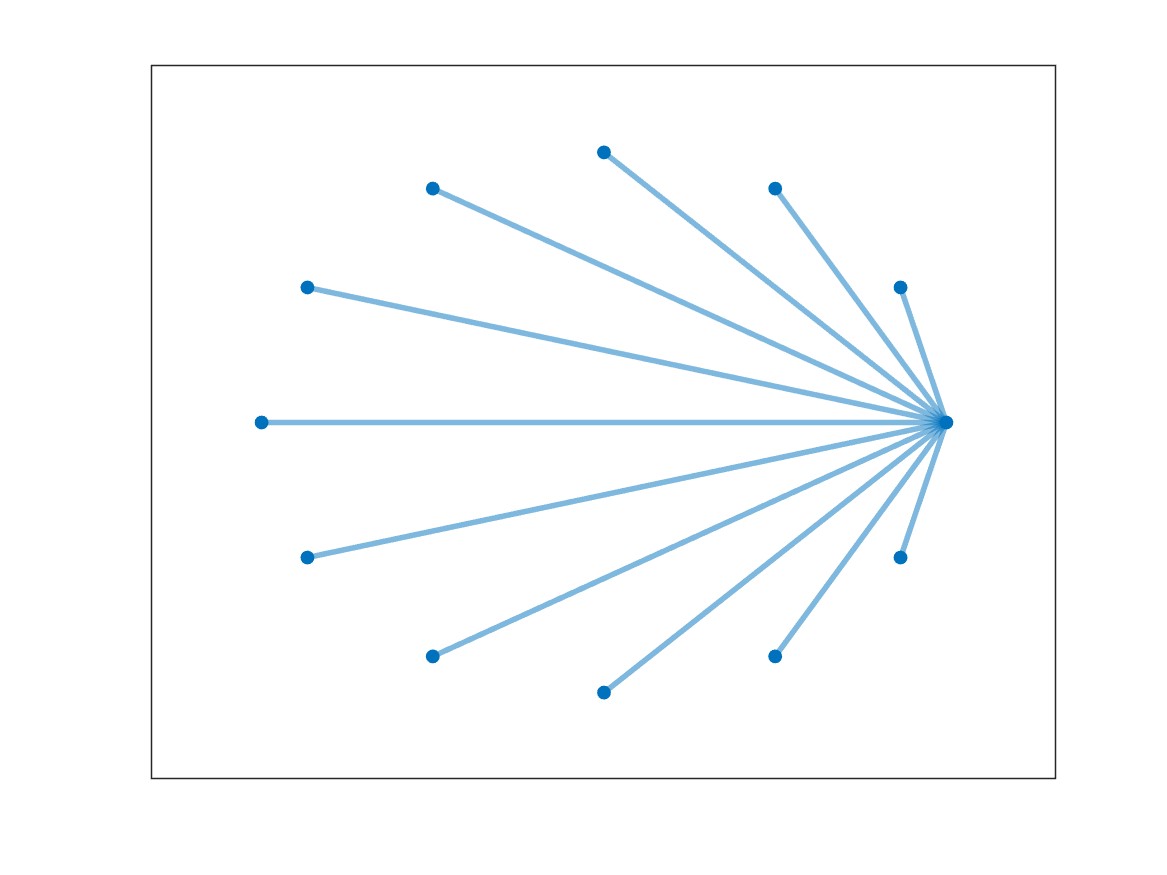}
\includegraphics[trim = 60mm 40mm 70mm 30mm,clip, width=6.2cm, height=4.2cm]{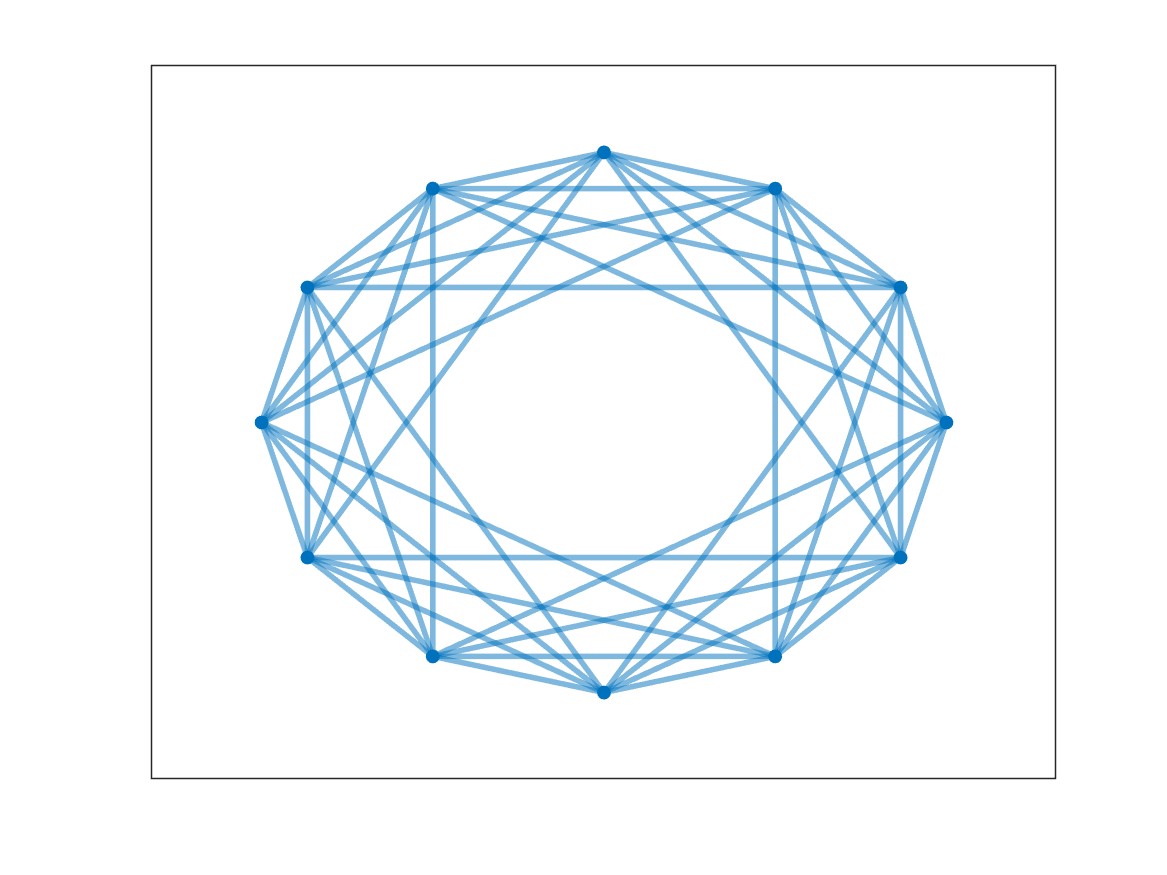}

(a) \hspace{4cm} (b)

\includegraphics[trim = 60mm 40mm 70mm 30mm,clip, width=6.2cm, height=4.2cm]{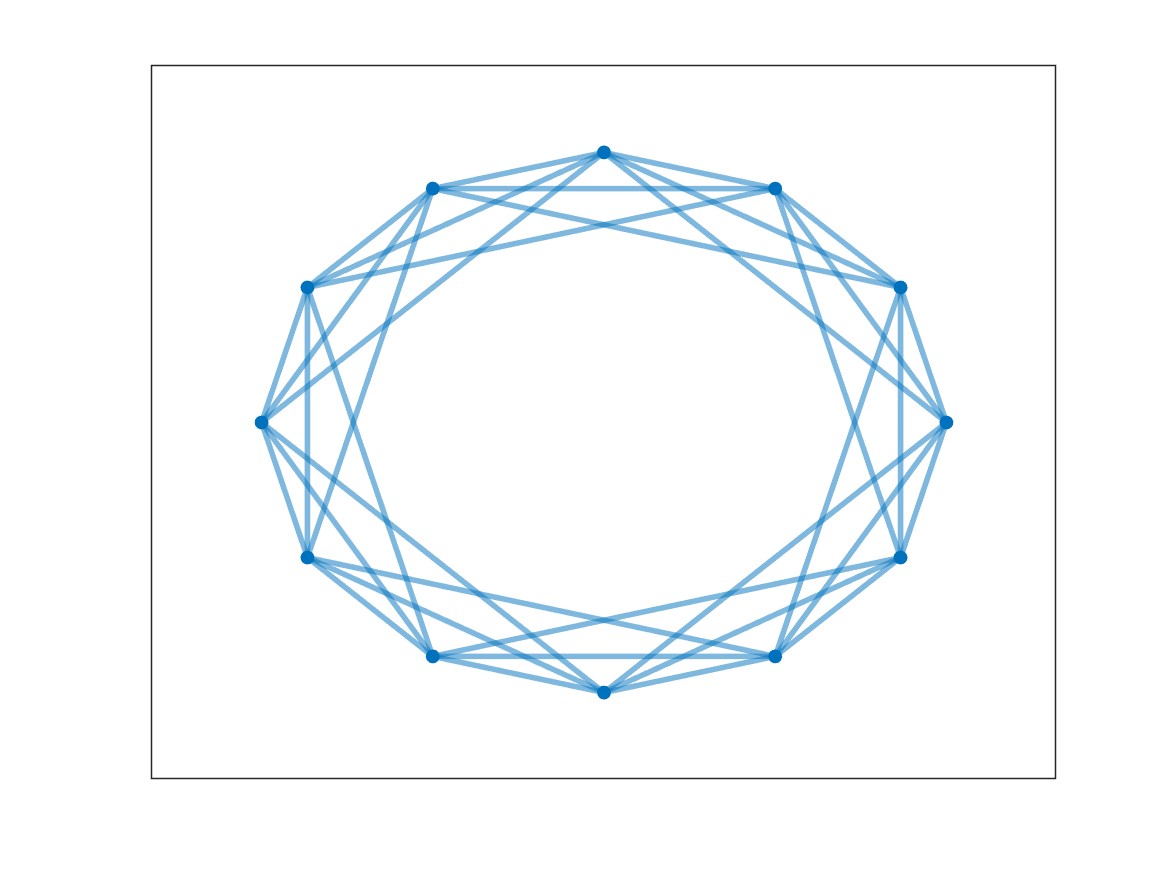}
\includegraphics[trim = 5mm 2mm 20mm 20mm,clip, width=6.2cm, height=4.2cm]{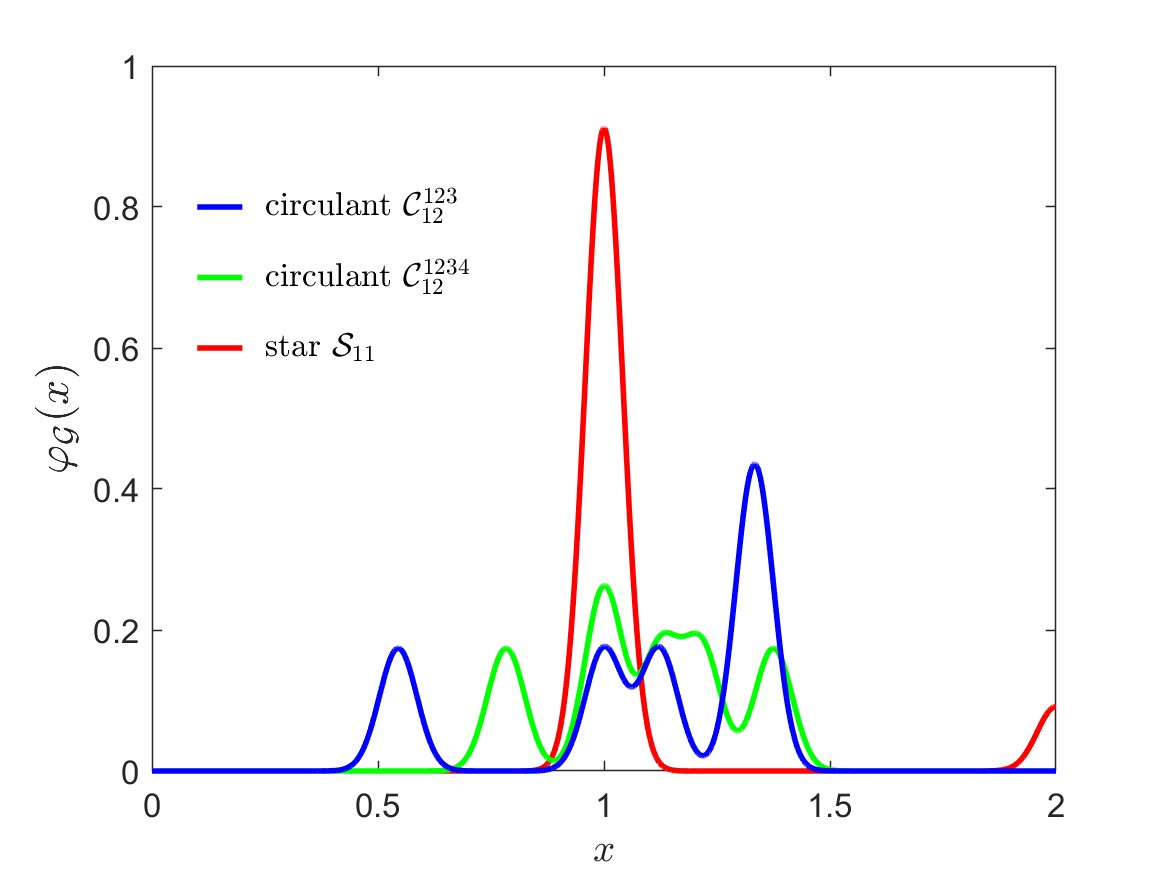}

(c) \hspace{4cm} (d)

\caption{Examples of target graphs and target densities for graph size $n=12$. (a) Star graph $\mathcal{S}_{11}$. (b) Circulant graph $\mathcal{C}_{12}^{1234}$. (c) Circulant graph $\mathcal{C}_{12}^{123}$. (d) Spectral densities of $\varphi_{\mathcal{T}}(x)$, see Eq. \eqref{eq:density}, of the target graphs (a)-(c).}
\label{fig_graphs}
\end{center}
\end{figure}

For each candidate graph  $\mathcal{G}$, we calculate the smoothed spectral density \eqref{eq:density} and  compare it to the spectrum of a target graph $\mathcal{T}$.  
Therefore, we use a pseudometric on graphs by the distance~\cite{gu16} 
\begin{equation}
d(\mathcal{G},\mathcal{T})= \int_0^2|\varphi_{\mathcal{G}}(x)-\varphi_{\mathcal{T}}(x)|dx \label{eq:distance}
 \end{equation}
 as the fitness function which should be minimized.
In the numerical experiments reported in Section \ref{sec:results}, we use  three classes of target graphs. We take star graphs $\mathcal{S}_{k}$ with size $n=k+1$, which have $k$ leaves connected to a central hub, and two types of circulant graphs $\mathcal{C}_n^L$ of size $n$, where each vertex $i$ is connected to the ($i+j$)-th and ($i-j$)-th vertex for each $j$ in a list $L$. We consider $L=1,2,\ldots,\lfloor{n/3}\rfloor$ and $L=1,2,\ldots,\lfloor{n/4}\rfloor$.  Fig. \ref{fig_graphs} shows the graphs $\mathcal{S}_{11}$, $\mathcal{C}_{12}^{1234}$ and $\mathcal{C}_{12}^{123}$   and their spectral densities $\varphi_{\mathcal{T}}(x)$ for $n=12$.

The algorithm proceeds over generations $\kappa=1,2,\ldots,K$. 
The performance of the algorithm is recorded by the final best fitness measured as the smallest distance 
$d(\mathcal{G},\mathcal{T})$ obtained after the final generation $K$.

\subsection{Genetic Operators} \label{sec:genetic}
In the following, we describe the mutation and crossover operators used in our approach.
\subsubsection{Mutation}
For unweighted graphs, the adjacency matrix used as representation only contains zeros (no edge between vertices) and ones (edge between vertices). New individuals can obtained by flipping zeros and ones in the adjacency matrix, which is equivalent to removing and adding edges~\cite{de18,ver17,loz21}. 
This serves as the basis of our mutation operator.
Mutation is carried out with a mutation rate $\alpha$, which implies that with a probability of $1-\alpha$ graphs are not altered by mutation.  The
mutation strength $\beta$ indicates the number of added (or removed) edges per graph and generation.  
If a graph is chosen to undergo mutation, edges can either be  added or removed. Given that the graphs are expected to evolve towards a target spectrum, preliminary experiments (not reported here due to brevity) have shown that, for performing properly, the balance between edge additions and edge removals  should be tuned based on structural characteristics of the evolving graph in relation to those of the target spectrum. Consider the case where the graph $\mathcal{T}$ with the target spectrum is a rather dense graph with a larger average degree, and large cliques and short paths, while the evolving graph $\mathcal{G}$ (or even an initial graph) is rather path-like with a smaller average degree, and long paths and small clusters. These structural differences are clearly reflected by a spectral difference between the Laplacian eigenvalues $\lambda(\mathcal{G})$ and $\lambda(\mathcal{T})$, which is  expressible by the distance $d(\mathcal{G},\mathcal{T})$, see Eq.~\eqref{eq:distance}. As in the considered case 
the evolutionary search is expected to guide the evolving graph in a transition from the spectrum of a path-like graph to the spectrum of a more dense graph, mutation should have more edges added than removed. If the evolving (or initial) graph is rather dense and the target graph is rather path-like, then mutation should prefer removing edges over adding edges.

A good single parameter for measuring the structure of a graph is the algebraic connectivity, which is the second smallest eigenvalue $\lambda_2$ of the (normalized) Laplacian. This quantity not only accounts for how well a graph is connected; it has also been shown that $\lambda_2$ is suitable to assess whether graphs are rather path-like or dense, have a rather smaller or larger average degree and possess rather short or long paths~\cite{deabr07,ban08,ban09,wills20,rich21}.  If $\lambda_2=0$, the graph is disconnected, low values of $\lambda_2$ suggest path-like graphs with smaller average degree and long paths, while higher values of $\lambda_2$ imply more dense graphs with larger average degree and shorter paths. 

Thus, and
according to the design principle of using spectral quantities, we suggest mutation control guided by the algebraic connectivity $\lambda_2$.  If a graph is chosen to undergo mutation, and its $\lambda_2$ is smaller than the algebraic connectivity of the target graph, then more edges are added.  If  $\lambda_2$ is larger than the algebraic connectivity of the target graph, then more edges are removed.
Furthermore, it has shown to be favorable if for edge addition
the selection of vertices undergoing mutation is slightly biased depending on the degree of one of the vertices involved. We discuss the effects of the two possibilities: either vertices with a high degree are favored, or vertices with a low degree are preferred. 
Adding edges to high degree vertices has some favorable structural consequences as long as the algebraic connectivity $\lambda_2$ is smaller than that of the target graph, but considerably larger than zero.   High degree vertices already act as structural hubs. Thus, adding edges to them can amplify their role, improving global connectivity. Furthermore, we noticed that  
when many vertices have low degree, adding edges to them distributes changes thinly across the graph.
Focusing on high degree vertices instead concentrates modifications where they have a stronger effect on spectral characteristics.

However, the effect should be reversed for $\lambda_2$ close to zero (or the graph having a low minimum degree $\underset{i}{\min} \: k_i$). Here, edge addition should be favored for low degree vertices, as favoring edge additions to them reduces the risk of really sparsely connected vertices. Moreover, removing an edge from a graph with low $\lambda_2$ would risk disconnecting the graph and thus rendering the mutation ineffective. The bias is implemented by selecting vertices at random with a probability directly (or inversely) depending on their degree.

The two-stage mutation control of edge addition is implemented as follows, see Algorithm 1 for details. If $\lambda_2$ of the evolving graph $\mathcal{G}$ is smaller than $\lambda_2(\mathcal{T})$ of the target graph and above a threshold $\lambda_2^t>0$ (and above a minimum degree $\underset{i}{\min} \: k_i$) 
two vertices which are not connected by an edge are selected at random. The random selection is biased to favor higher degrees of one of the selected vertices.   An edge is inserted between the selected vertices. 

If $\lambda_2$ is below the threshold, the same is done with a lower degree favored.
If  $\lambda_2$ is larger than the algebraic connectivity of the target graph, edges are removed. Two vertices which are connected by an edge  are selected uniformly at random and the edge is deleted. Preliminary experiments (not reported here) have shown that for edge removal biased selection of vertices is not generally improving performance. Furthermore, it is checked if the edge removal disconnects the graph. If this is the case, the edge removal is taken back and mutation repeated.

\begin{algorithm}[t]
\SetAlgoLined
\SetKwInOut{Input}{Input}\SetKwInOut{Output}{Output}
 \Input{ Population of graphs $P_\mathcal{G}$, target graph $\mathcal{T}$, mutation rate $\alpha$; mutation strength $\beta$, threshold $\lambda_2^t$}
 \Output{ Population of mutated graphs %$P_{\mathcal{G}_m}$
 ;}
Calculate $\lambda_2$ of the target graph\;
 \For{each graph $\mathcal{G}$ in $P_\mathcal{G}$ }
 {
 Proceed with a probability $\alpha$\;
 Calculate $\lambda_2(\mathcal{G})$\;
  \Repeat {$\beta$ times carried out } {
  \eIf{$\lambda_2(\mathcal{G}) \leq \lambda_2(\mathcal{T})$} 
{\eIf{$(\lambda_2(\mathcal{G})>\lambda_2^t)$ $\land$ ($\underset{i}{\min} \: k_i<2$) }
{
 
 Select two not adjacent vertices at random\;
 Bias one vertex directly with degree\; Connect the vertices by an edge\; } 
{ Select two not adjacent vertices at random\;
 Bias one vertex inversely with degree\; Connect the vertices by an edge\;}
}
{Select two adjacent vertices uniformly at random \; Delete edge between them\;
Check if graph is still connected, if not, undo mutation and repeat\;}

}      
 }

  \caption{Mutation control: edge addition and edge removal }
\end{algorithm}

\subsubsection{Crossover}
While mutation by edge removal or edge addition changes the graph on a comparably small scale and only locally on the level of individual vertices, crossover intends to alter the graph structure on a more global level. Such a change in the global structure can be achieved by applying principles of crossover to graphs, which generally means selecting parent graphs, identifying (favorable) inheritable traits, and combining them to produce offspring.  For two parent graphs, this may mean cutting each of them into two (or even more) subgraphs and exchanging the (compatibly sized) cuts. Thus, a rather straightforward implementation of graph crossover, which we call  basic crossover, involves (after selecting two graphs) dividing each graph  into two subgraphs by randomly selecting subsets of vertices~\cite{de18,glo00}, see Algorithm 2 for details. We use the same random subset size for both parent graphs. Because we do not want to have very small and trivial subgraphs, a minimum subset size $\nu=3$ is observed. The edges among the vertices of each subgraph are preserved while the edges between the subgraphs are discarded. As the parent graphs are cut with compatible subset size, we have two subgraphs each that can be swapped and glued together by adding randomly edges between them.  We add as many edges as have been deleted by the subgraph partition.   

\begin{algorithm}[t]
\SetAlgoLined
\SetKwInOut{Input}{Input}\SetKwInOut{Output}{Output}
 \Input{Population of graphs $P_\mathcal{G}$, minimal subgraph size $\nu$}
 \Output{ Population of crossover graphs %$P_{\mathcal{G}_m}$
 ;}

 \For{$i=1:\ell/2$}
 { Select two graph uniformly at random from $P_\mathcal{G}$\; 
 Select cut point from a uniform random variable distrubuted on $[\nu,n-\nu]$\;
 Cut both graphs at cut point \; Delete edges between the cuts\; Store number of edges deleted\;
 Rearrange graph cuts to obtain two graphs of size $n$\;
 Add randomly same number as deleted edges between cuts\;
  }

  \caption{Basic graph crossover (bc)}
\end{algorithm}

\begin{algorithm}[t]
\SetAlgoLined
\SetKwInOut{Input}{Input}\SetKwInOut{Output}{Output}
 \Input{Population of graphs $P_\mathcal{G}$}
 \Output{ Population of crossover graphs %$P_{\mathcal{G}_m}$
 ;}

 \For{$i=1:\ell$}
 { Select graph uniformly at random from $P_\mathcal{G}$\; 
Calculate Fiedler vector and cut graph according to spectral clustering with 2 clusters\;
 Delete edges between the cuts\; Store number of edges deleted\;
 Add randomly same number as deleted edges between cuts\;
  }

  \caption{Spectral crossover 1 (sc1)}
\end{algorithm}

Although such an implementation is viable, there are also some disadvantages~\cite{glo00,kim11}. Randomly cutting or partitioning a graph may disrupt important topological or functional properties, such as connectivity, cycles, or clustered communities, resulting in offspring that possibly lacks meaningful features inherited  from their parents. This may hinder useful traits and high-quality building blocks from propagating. Because the partition is independent of the graph structure, the likelihood of producing offspring with improved fitness is possibly lower than in crossover designed to respect domain-specific (e.g., community, motif) structures, for instance  multiplicity constraints~\cite{th22} or motif preservation~\cite{mur24}. In order to address these issues, and consistent with our design principles,  we study two crossover schemes which use spectral techniques~\cite{martin06,kim11}. In detail, we use for subgraph partitioning a spectral clustering with two clusters, which is based on the Fiedler vector of the normalized Laplacian $\Lambda_{\mathcal{G}}$~\cite{shi00,hig07,jia14}. The Fiedler vector is the eigenvector corresponding to the second smallest eigenvalue $\lambda_2$ of $\Lambda_{\mathcal{G}}$ and its signs separate vertices into two distinct clusters. The division obtained is such that connections across the groups are minimized, resulting in a partition that often reflects the main components of the graph structure. 
In other words, the partition reflects the underlying cluster structure without being dominated by high degree vertices. 

\begin{algorithm}[t]
\SetAlgoLined
\SetKwInOut{Input}{Input}\SetKwInOut{Output}{Output}
 \Input{Population of graphs $P_\mathcal{G}$}
 \Output{ Population of crossover graphs %$P_{\mathcal{G}_m}$
 ;}

 \For{$i=1:\ell$}
 {  
Calculate Fiedler vector and cut graph into subgraphs according to spectral clustering with 2 clusters\;
 Delete edges between the cuts\; Store number of edges deleted\;
 Store complementary subgraphs according to  size group\; 
   }
 \For{Each size of size group}
 {
Select randomly a pair of complementary subgraphs\;
Recombine the pair to get a graph of size $n$ \;
Connect the graph by randomly adding edges\;

 }

  \caption{Spectral crossover 2 (sc2)}
\end{algorithm}

We compare two implementations: A basic one (spectral crossover 1, see Algorithm 3) where only one parent graph is used, which is divided according to spectral clustering with two clusters. The edges within each cluster are preserved, the edges between the clusters are deleted and replaced by the same number of randomly inserted edges.  A second implementation (spectral crossover 2, see Algorithm 4) first carries out a spectral clustering with two clusters for all graphs of the population. By cuts according to the clustering, we obtain  pairs of parental subgraphs with complementary size. The cuts remove the edges between the subgraphs, while the edges within subgraphs are preserved. If for instance, a set of graphs of even size $n$ (a population) undergoes spectral clustering with two clusters, then some graph may be clustered into subgraphs of size $n/2$  and $n/2$, some others into subgraph of size $n/2-1$ and $n/2+1$, and even others with size $n/2-2$ and $n/2+2$, and so on. Thus, we get the size groups $[n/2]$, $[n/2\pm 1]$, $[n/2\pm 2]$, and so on.   We sort and store the complementary subgraphs according to their size group. When this is done for all graphs of the population, we take the complementary subgraphs from each size group and randomly recombine them to rearrange graphs of size $n$. We connect the subgraphs by randomly adding edges between them.   The subgraph recombination typically involves cuts from  different parental graphs, and thus different numbers of edges have been deleted. We randomly add as many edges as the (rounded) average.

The degree of mixing between parental subgraph structures produced by this spectral crossover is inherently constrained by the number of distinct subgraphs available within each size group. In principle, the size group can be as small as to contain just one pair of subgraphs. Thus, the parental subgraphs would be reconnected to themselves, which would mean that  for a size group with just one pair spectral crossover 2 would coincide with spectral crossover 1. However, experiments have shown that for sufficient population size and diversity, size groups with just one pair is very rare.

\begin{figure}[t]

\includegraphics[trim = 10mm 11mm 10mm 75mm,clip, width=8.0cm, height=4.8cm]{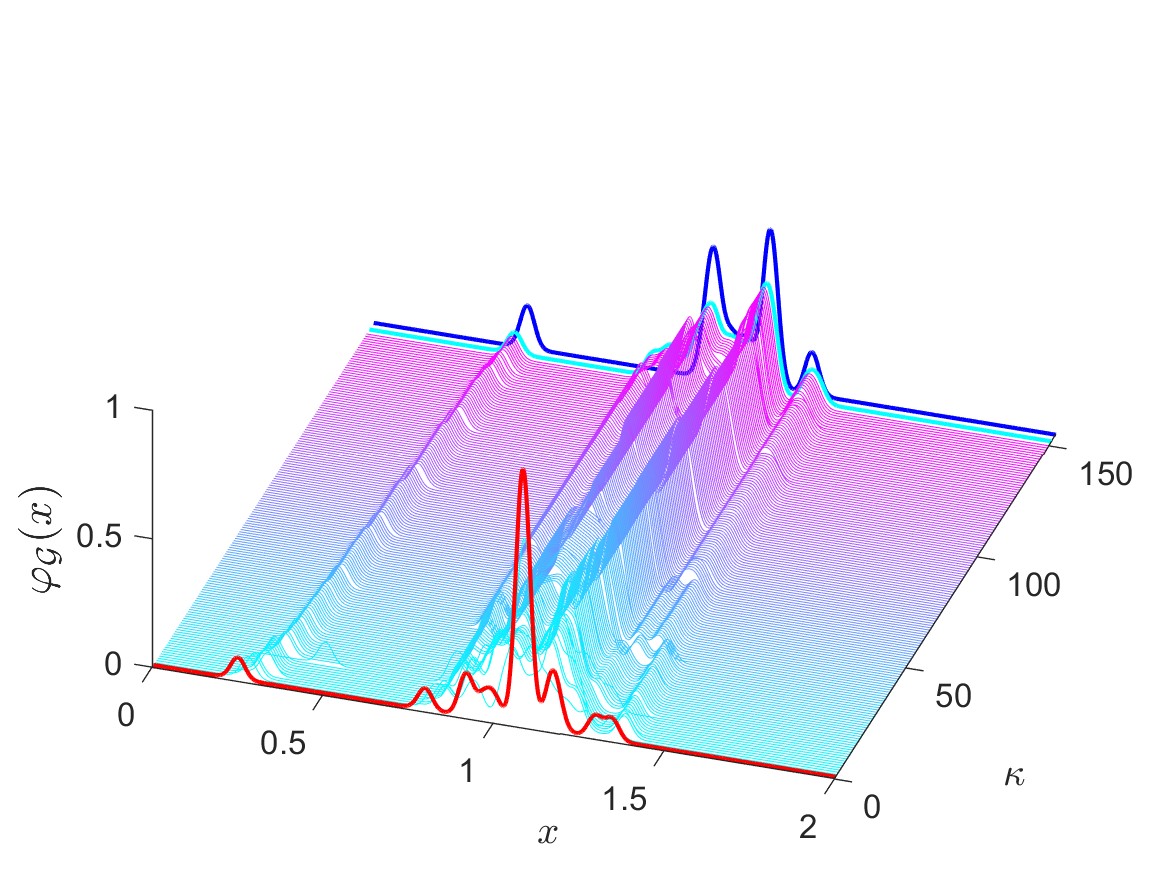}

\caption{Evolution of the spectral density $\varphi_{\mathcal{G}}(x)$ over generations $\kappa$ for $n=24$, $12$-regular initial graphs and the target graph $\mathcal{C}_{24}^{1,2,\ldots,6}$. }
\label{fig_dens_gene}
\end{figure}

\section{Experimental Investigations} \label{sec:results}
We now carry out experimental investigations for generating graphs using our approach 
for evolving graphs towards desired Laplacian spectra. We consider the generation of different types of graphs and graph size, and the diversity of the generated graphs, see Tab. \ref{tab:map} for the algorithm's design parameters.

\begin{table}
\caption{Design parameter of the algorithm.}
\begin{tabular}{lll}  Design parameter & Symbol & Value(s) \\
\hline \hline  Population size & $\ell$ & 40 \\ 
Generations &$K$&1000 \\
Tournament size &$s_t$& 2\\
Mutation rate & $\alpha$ & 0.75\\
Mutation strength & $\beta$ & 4\\
Connectivity threshold& $\lambda_2^t$ & 0.001\\
Minimal subgraph size& $\nu$&3\\ \hline
Crossover & & bc, sc1, sc2\\
\hline Graph size & $n$ & 24, 64, 128, 256, 512 \\ \hline
 \hline
    \end{tabular}
\label{tab:map}
     \end{table}

\subsection{Results for Different Target Graphs}

As an illustration of the evolution of the spectral density $\varphi_{\mathcal{G}}(x)$, Fig.~\ref{fig_dens_gene} shows  the first 150 generations for $n=24$, $12$-regular initial graphs and the target graph $\mathcal{C}_{24}^{1,2,\ldots,6}$. The red line for $\kappa=0$ is the density of the best individual from the initial population. The blue line at $\kappa=150$ is the density of the target graph. We see an approximate transition towards the target density.      

\begin{figure}[h!]
\includegraphics[trim = 0mm 0mm 0mm 0mm,clip, width=8.0cm, height=4.8cm]{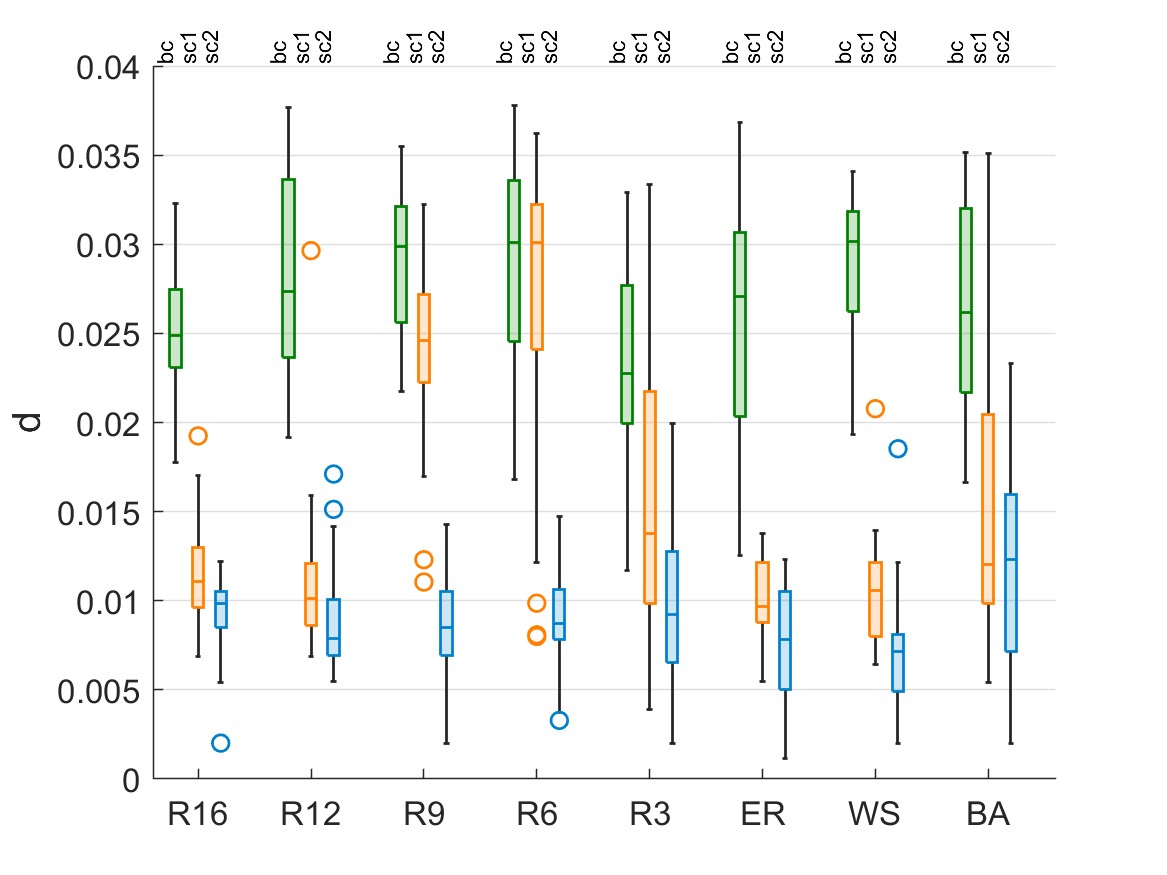}

(a)

\includegraphics[trim = 0mm 0mm 0mm 0mm,clip, width=8.0cm, height=4.8cm]{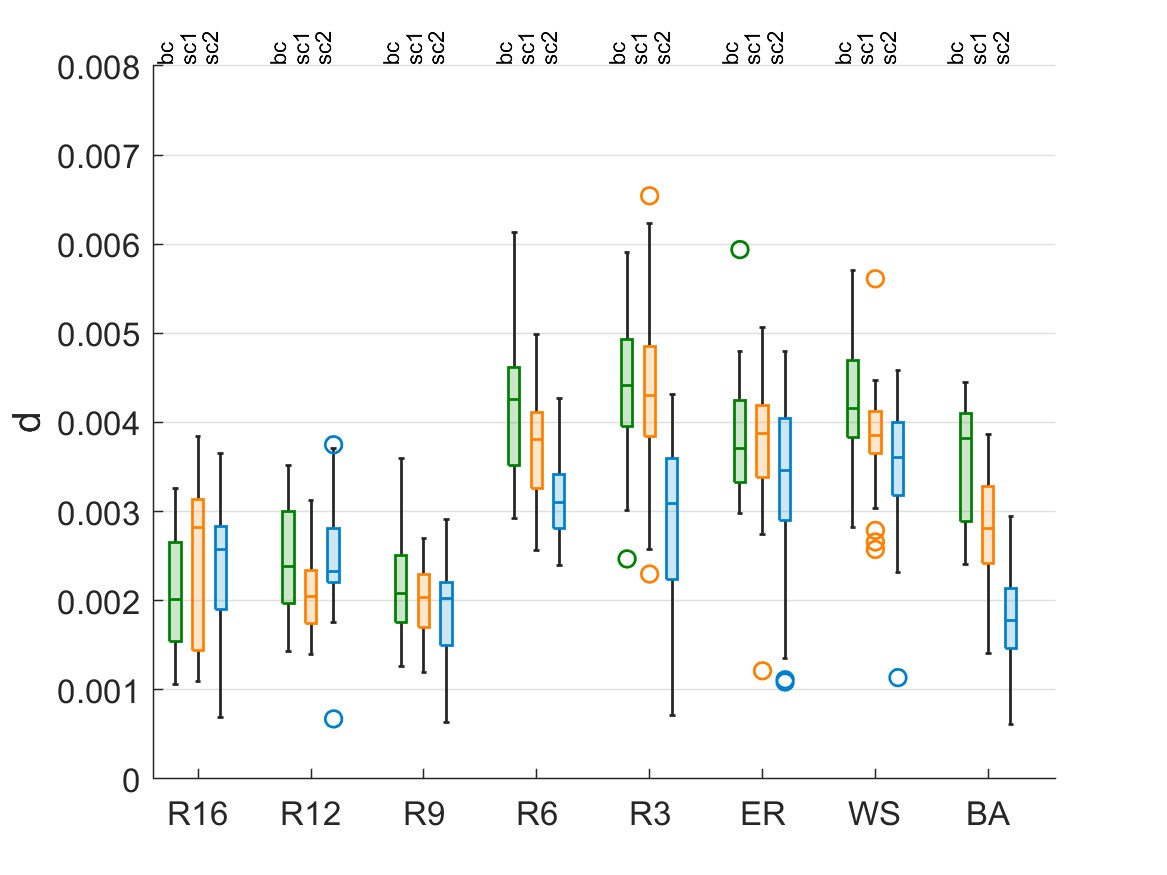}

(b) 

\includegraphics[trim = 0mm 0mm 0mm 0mm,clip, width=8.0cm, height=4.8cm]{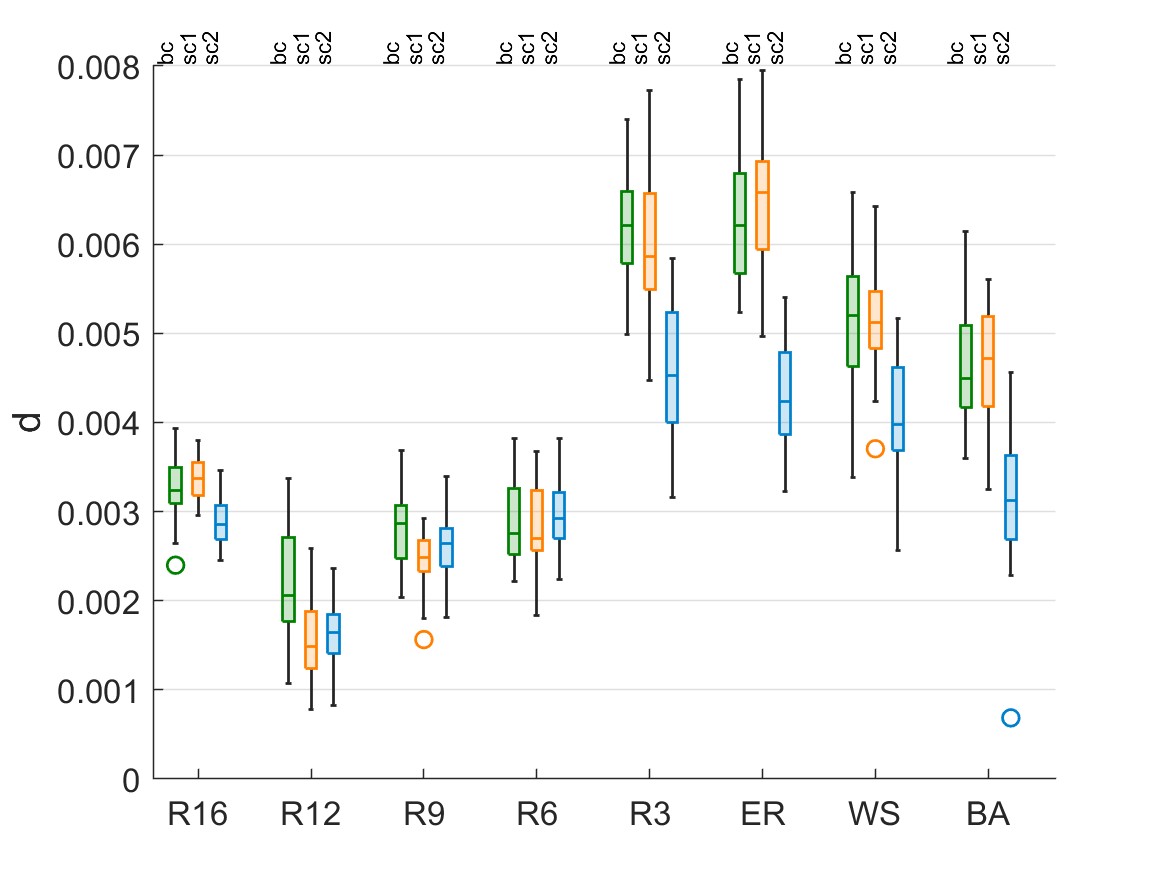}

(c) 
\caption{Results for target graphs of size $n=24$. Boxplots for 30 runs and 1000 generations showing the final best fitness measured by the distance $d$, see Eq. \eqref{eq:distance}, for eight initial graph populations and three crossover implementations. (a) Star graph $\mathcal{S}_{23}$ (b) Circulant graph $\mathcal{C}_{24}^{1,2,\ldots,8}$. (c) Circulant graph  $\mathcal{C}_{24}^{1,2,\ldots,6}$.  }
\label{fig_results}
\end{figure}

Fig.~\ref{fig_results} shows boxplots of the final best fitness measured by the distance $d$, see Eq. \eqref{eq:distance}, for 30 runs with 1000 generation each. We use three target graphs $\mathcal{T}$ with graph size $n=24$. The target graphs are: Fig.~\ref{fig_results}(a), star graph $\mathcal{S}_{23}$, Fig.~\ref{fig_results}(b), circulant graph $\mathcal{C}_{24}^{1,2,\ldots,8}$,  and Fig.~\ref{fig_results}(c), circulant graph  $\mathcal{C}_{24}^{1,2,\ldots,6}$.
 The results are for 8 initial graph populations:  pair-wise nonisomorphic $k$-regular graphs for $k=\{16,12,9,6,3\}$ (R16-R3), and   random graphs,  Erd\"os--Renyi (ER) with $p=0.3$, Barabasi-Albert (BA) with $m_0=8$ and $m=5$ and Watts-Strogatz (WS)  with $K=4$ and $\beta=0.3$. We consider
 three crossover implementations: basic graph crossover (bc), spectral crossover 1 (sc1) and spectral crossover 2 (sc2), see Algorithm 2-4. 

 We observe notable performance differences between different initial populations and also between different crossover implementations. Overall, the findings demonstrate that spectral crossover 2, Algorithm 4, frequently outperforms the other two implementations, in many cases by substantial amounts,  particularly for the star graph target. These results are in favor for the assumption that a partitioning which randomly disrupts structural features of the graph is less likely to support evolutionary search.

 The results for the circulant graph targets are considerable better than for the star graph target. Also, the differences between different initial populations are much clearer. 
 While for the star graph, Fig.~\ref{fig_results}(a), performance differences are not particularly substantial, they are much more prominent for the circulant graphs, Fig.~\ref{fig_results}(b) and (c). Here, initial populations of $k$-regular graphs give better results, with $k=\{16,12,9\}$ being beneficial for both   $\mathcal{C}_{24}^{1,2,\ldots,8}$. and  $\mathcal{C}_{24}^{1,2,\ldots,6}$. Regular initial graphs with $k=6$ are an ambiguous  case, given that they produce strong results for  $\mathcal{C}_{24}^{1,2,\ldots,6}$, but weaker performance for $\mathcal{C}_{24}^{1,2,\ldots,8}$. Regular initial graphs with $k=3$ and random initial graphs (with some less distinction for BA graphs) work less well for both circulant target graphs. 

A main difference between the two types of circulant graphs is  their density, see Fig.~\ref{fig_graphs}(b) and (c) for the examples $\mathcal{C}_{12}^{1,2,3,4}$ and $\mathcal{C}_{12}^{1,2,3}$. Thus, the  circulant graph $\mathcal{C}_{24}^{1,2,\ldots,8}$ is more dense (and has degree 16) than $\mathcal{C}_{24}^{1,2,\ldots,6}$ (which has degree 12). In the implementation discussed in this paper, evolutionary search for graph structures   acts solely by edge manipulations. 
As already discussed above, the extent to which such edge manipulations necessitate predominantly edge removals or edge additions (but also edge rewiring by alternating edge removal and edge addition) is also governed by structural differences between the initial graphs and the target graph.  The graph $\mathcal{C}_{24}^{1,2,\ldots,8}$ is denser than $\mathcal{C}_{24}^{1,2,\ldots,6}$, and both are even denser than $\mathcal{S}_{23}$.
 Thus, for $k$-regular initial graphs with $k=\{16,12,9\}$ and both circulant target graphs a lesser amount of edges must be added and edges must mainly be rewired. For $k=3$ and both circulant target graphs, a more substantial amount of edges must be added. For $k=6$, the amount of edge addition seems more manageable for the less dense target graph $\mathcal{C}_{24}^{1,2,\ldots,6}$, but more problematic for the more dense $\mathcal{C}_{24}^{1,2,\ldots,8}$. For the random initial graphs the situation is somewhat related. Here, we find the expected average degree for ER graphs as $p(n-1)=6.9$, for WS graphs as $2K=8$ and for BA graphs as $2m=10$. However, these are expected values of a degree distribution for the initial graphs. This leads, on the one hand, to a broader variety of outcomes over runs, which can be seen by the larger interquartile range as well as the length of the whiskers. 
Moreover, we still occasionally get good results, particularly for BA initial graphs, which have an expected average degree of $2m=10$, which is similar in order to $k=\{16,12,9\}$.

\begin{figure}[h!]

\includegraphics[trim = 1mm 1mm 1mm 1mm,clip, width=8.0cm, height=4.8cm]{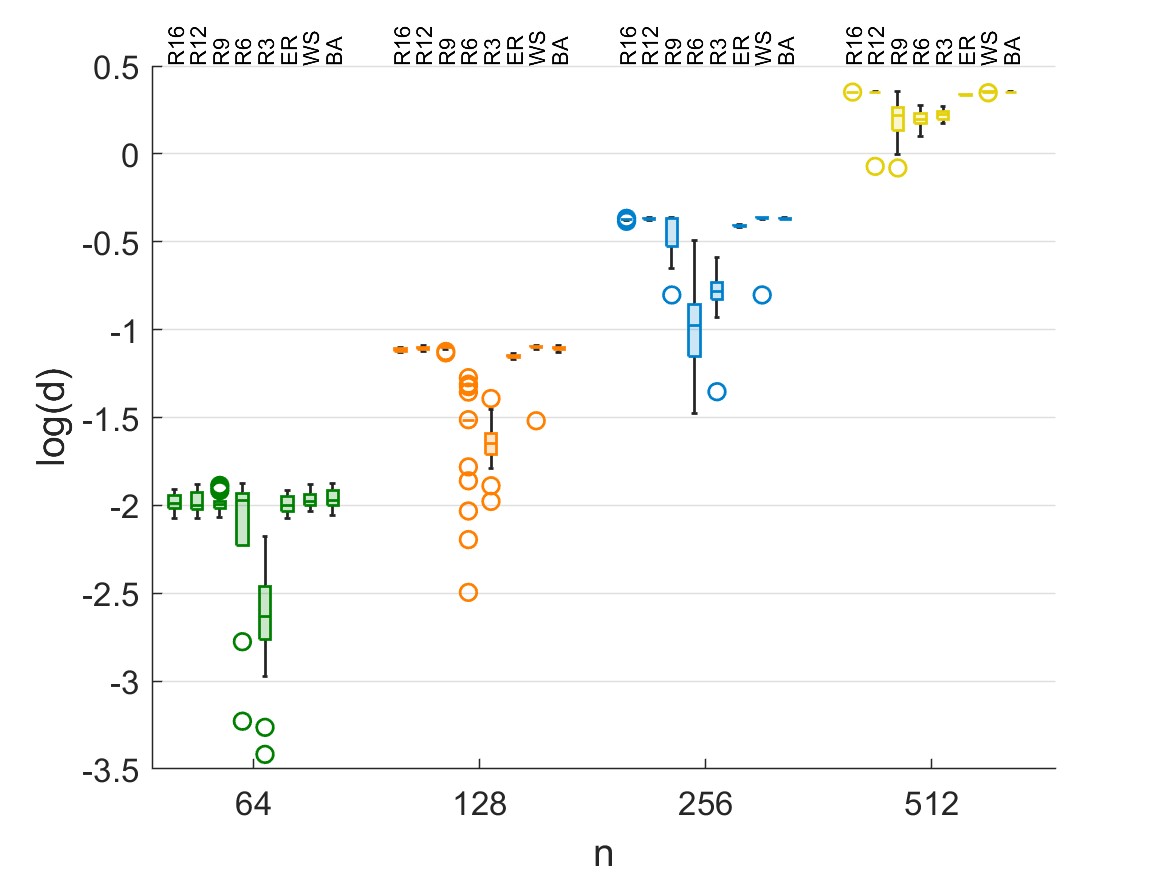}

(a)

\includegraphics[trim = 1mm 1mm 1mm 1mm,clip, width=8.0cm, height=4.8cm]{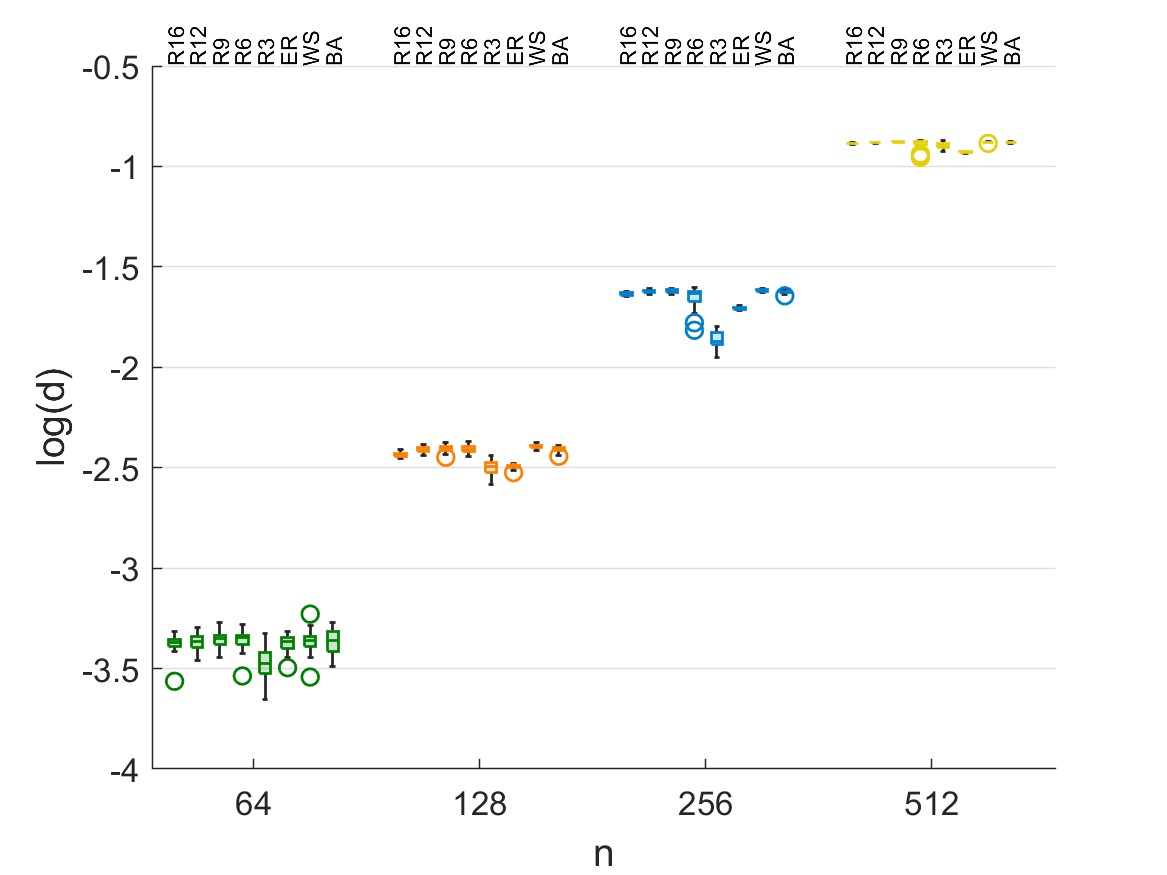}

(b) 

\includegraphics[trim = 1mm 1mm 1mm 1mm,clip, width=8.0cm, height=4.8cm]{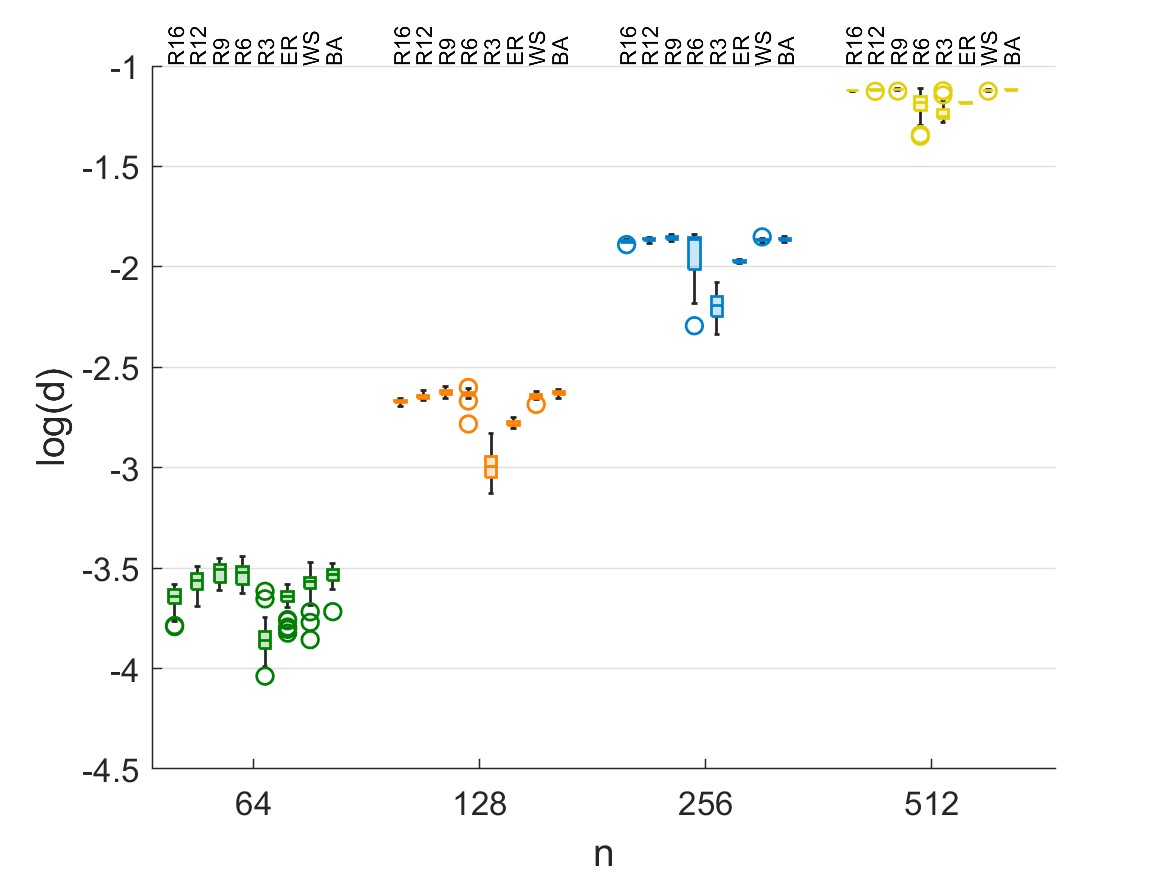}

(c)

\caption{Results for target graphs of varying size. Boxplots showing the final best fitness measured by the logarithmic distance $\log(d)$ for eight initial graph populations and graph size $n=\{64, 128, 256, 512\}$. (a) Star graph $\mathcal{S}_{n-1}$ (b) Circulant graph $\mathcal{C}_{n}^{1,2,\ldots,\lfloor  n/3\rfloor}$. (c) Circulant graph  $\mathcal{C}_{n}^{1,2,\ldots,\lfloor  n/4\rfloor}$. }
\label{fig_results2}

\end{figure}

These results for the circulant graph targets are contrasted by the star graph target. The star graph's density is very low, with all vertices but one having degree 1, and the hub having degree $n-1$. This yields an (average) degree of $2(n-1)/n=1.91$.   Yet, even very dense initial graphs do not perform substantially worse than less dense one.  Also,  for the random initial graphs, we find a larger range in results. Mainly edge removal is required for all initial graphs. Thus,    these results suggest a general feature: adding edges to foster evolutionary search is more subtile and sensitive to algorithmic details than removing edges, which we similarly also encountered with mutation.

\subsection{Results for Different Graph Sizes}

Fig. \ref{fig_results2} shows performance for varying size of the target graph.  We consider $n=\{64, 128, 256,512\}$, and the target graphs  $\mathcal{S}_{n-1}$,  Fig. \ref{fig_results2}(a),   $\mathcal{C}_{n}^{1,2,\ldots,\lfloor n/3 \rfloor}$,  Fig.~\ref{fig_results2}(b) and  $\mathcal{C}_{n}^{1,2,\ldots,\lfloor n/4 \rfloor}$,  Fig. \ref{fig_results2}(c). Only spectral crossover 2 is used. Note that the performance measured by the distance $d$, see Eq. \eqref{eq:distance}, is now on logarithmic scale. Unsurprisingly, the final best fitness generally gets larger as the graph size increases. Given that we use the same population size ($\ell=40$), it is, however, notable that fitness deteriorates gradually  and there are mostly no dramatic jumps in performance. Because both scales are logarithmic, the observed results suggest an approximate power-law relation between distance $d$ and graph size $n$ for a constant population size. In addition, there are some details worth discussing. First of all, as for $n=24$, the performance for the two circulant graphs is better than for the star graph. However, for these target graphs, the advantage of regular initial graphs with  $k=\{16,12,9\}$ is no longer present. This is in line with the observation discussed above for $n=24$ that evolutionary search is easier if fewer edges must be added. As for larger graph size $n$ also the degree of the circulant target graphs increases (the degrees are   2$\lfloor n/3 \rfloor$ and $2\lfloor n/4 \rfloor$, respectively), differences in the amount of required edge additions between the regular initial graphs employed  cease to make much of a difference.    
% For circulant graphs targets, we see that initial populations with $16$-regular graphs perform particularly well, very clearly for $n=72$. For $n=48$, also $12$-regular initial graphs give  performance advantages. This is in line with the observation discussed above for $n=24$.  Evolutionary search is easier if fewer edges must be added.  Also, we again observe ambiguous cases where $12$-regular graphs yield good results for $n=48$, but cease to do so for $n=72$.    For $n=48$, the circulant target graph has degree 24, while for $n=72$, it has degree 36, which might make it less manageable to add the required edges.
 
For the star graph target we find that results for the initial population with $k$-regular graphs with $k=\{6,3\}$ are considerably better than for all other initial populations, although for $n=512$ also $9$-regular graphs initial graphs work well. These performance differences are much clearer than for $n=24$.  This means for higher dimensions of evolving graphs, we also find for a low target density the effect that similarity between initial and target density seems helpful. An analysis has shown that this is not a convergence issue addressable by even more generations, but most likely a lack of graph diversity. Further works is needed to clarify the behavior.      

\subsection{Results for Different Graph Metrics}

A major application domain of evolving graphs towards Laplacian spectra is generating sets of graphs for experimentally evaluating network algorithms and protocols. Thus, while the generated graphs should match the target spectrum as closely as possible, they should not be identical or very similar  to the target graph itself with respect to other (particularly non-spectral) graph measures. 
\begin{figure}[h!]

\includegraphics[trim = 11mm 3mm 10mm 8mm,clip, width=6.2cm, height=4.2cm]{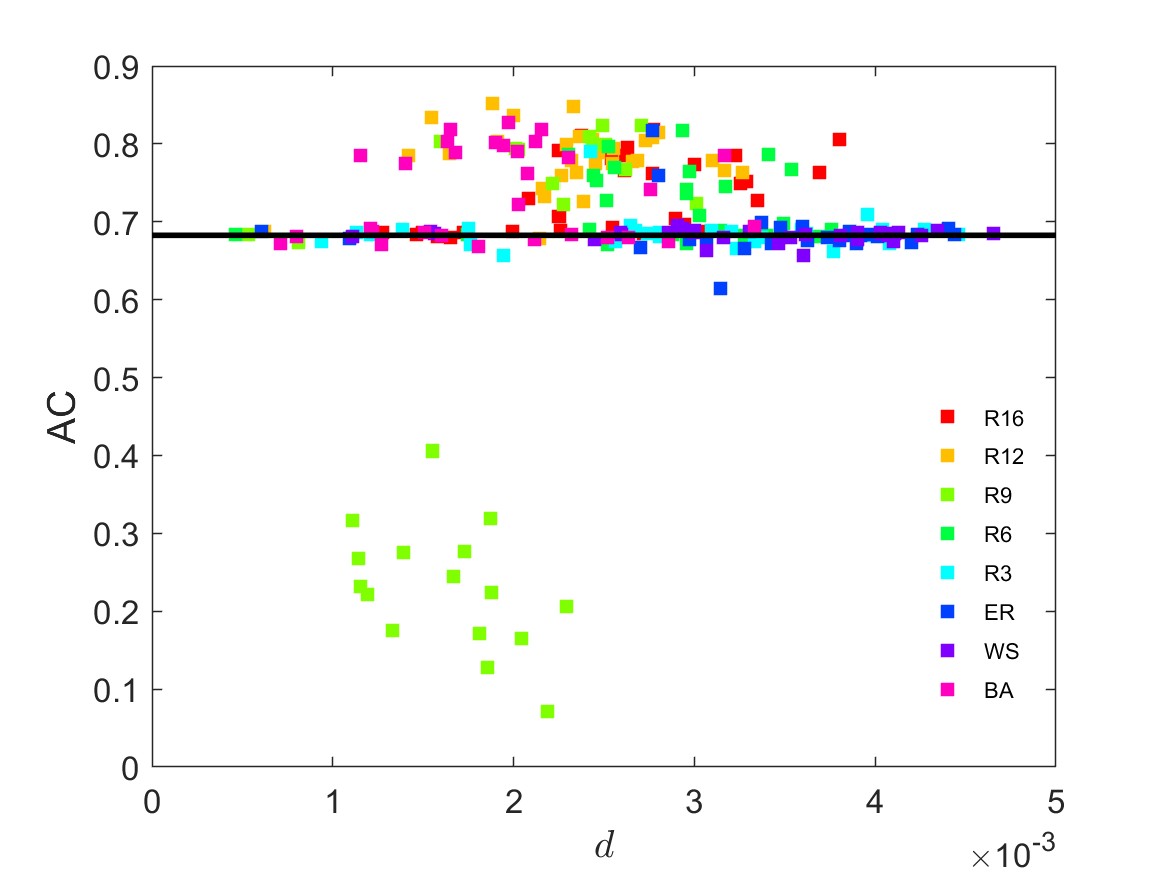}
\includegraphics[trim = 11mm 3mm 10mm 8mm,clip, width=6.2cm, height=4.2cm]{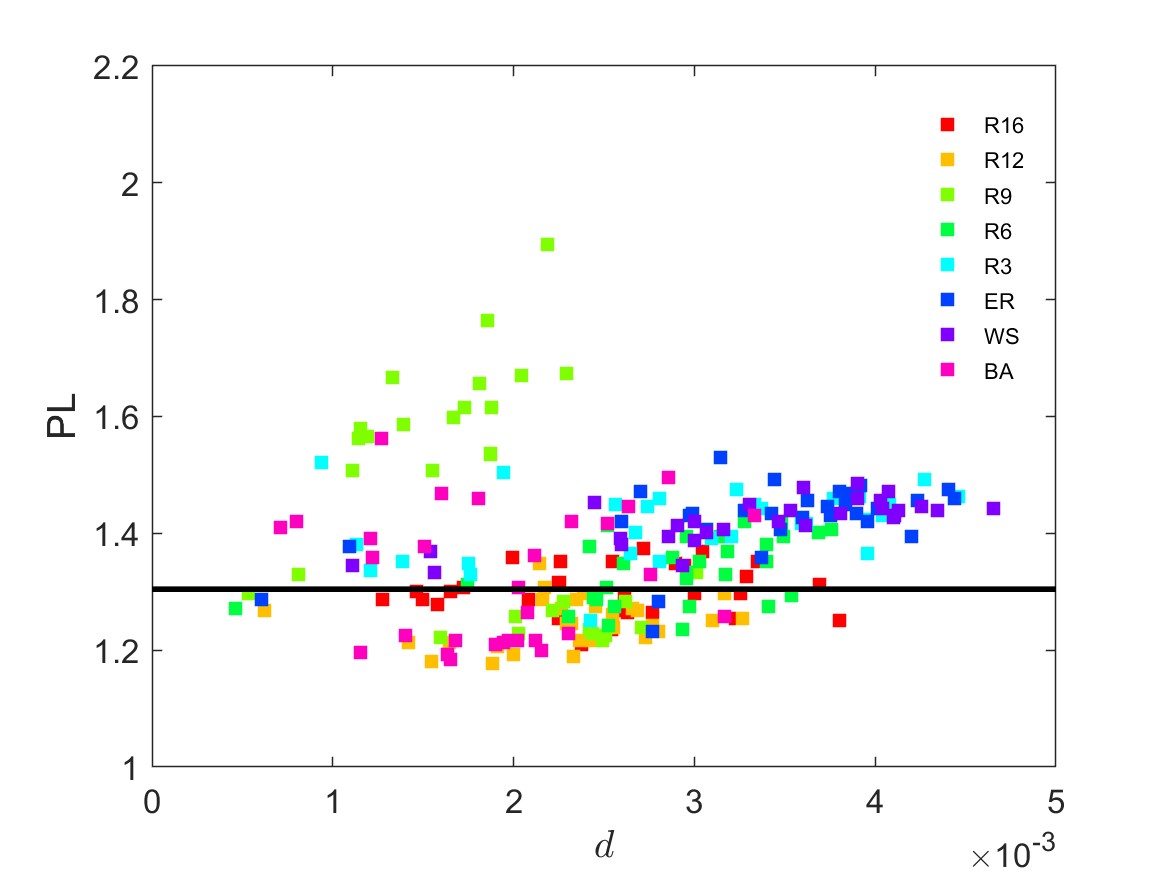}

(a) \hspace{7cm} (b)

\includegraphics[trim = 11mm 3mm 10mm 8mm,clip, width=6.2cm, height=4.2cm]{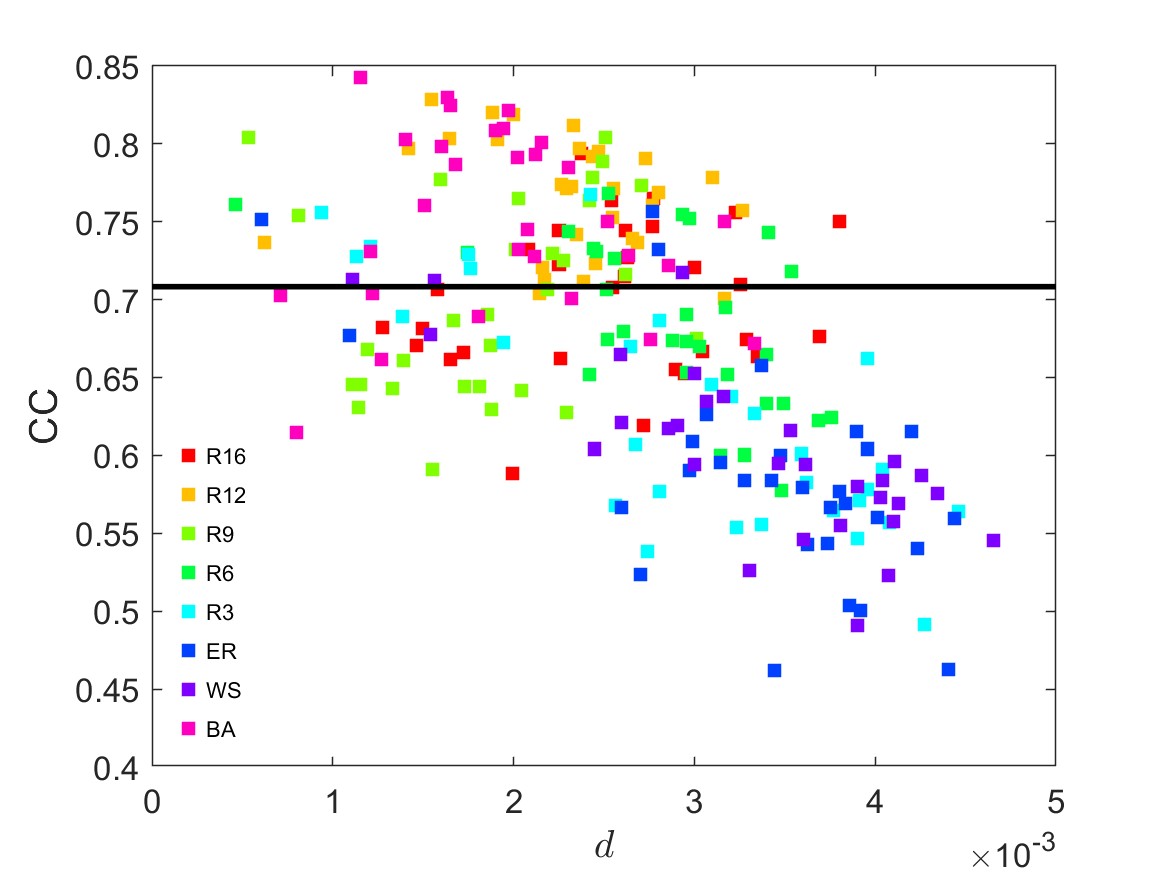}
\includegraphics[trim = 11mm 3mm 10mm 8mm,clip, width=6.2cm, height=4.2cm]{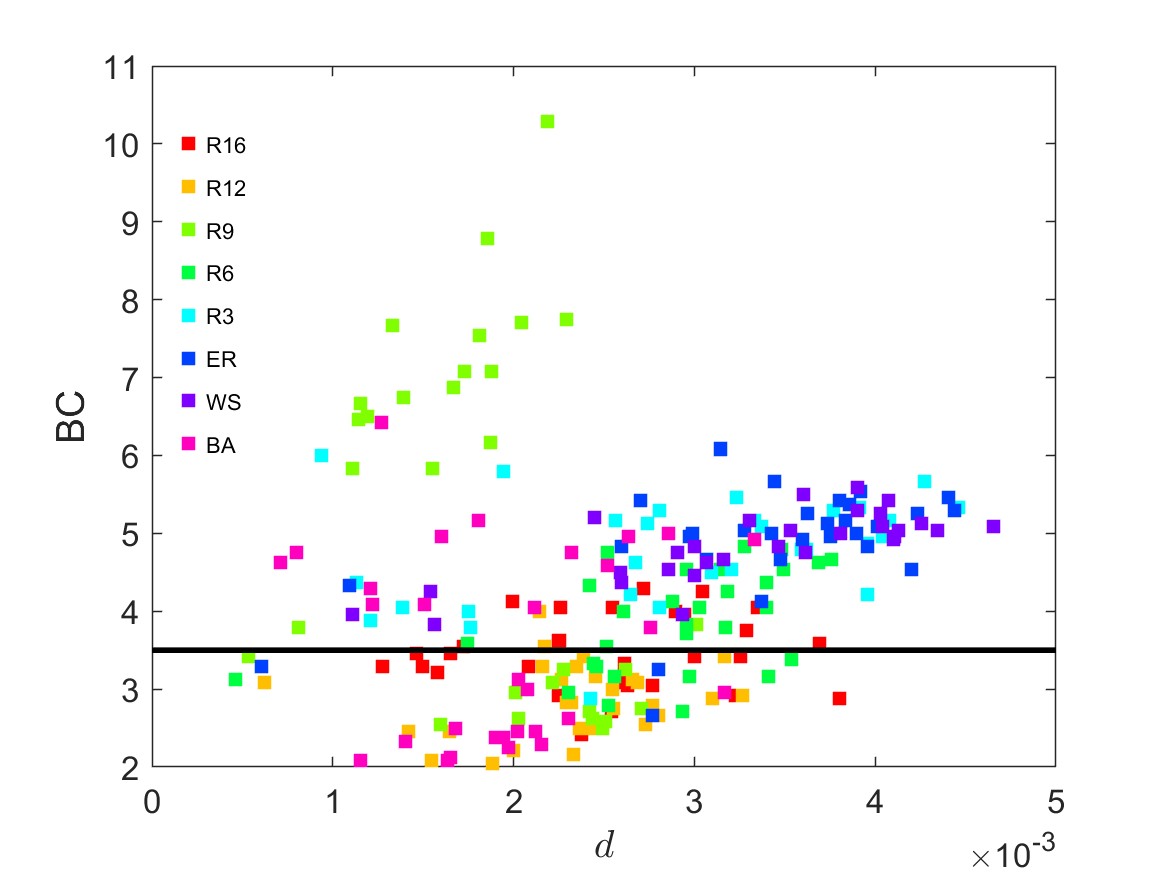}

 (c) \hspace{7cm} (d)
\caption{Scatter plots of final best fitness versus graph metrics for the circulant target graph $\mathcal{C}_{24}^{1,2,\ldots,8}$. The color code indicates different initial populations. (a) Algebraic connectivity (AC). (b) (Average) path length (PL). (c)  (Global) clustering coefficient (CC). (d) (Average) betweenness centrality (BC). }
\label{fig_results3}

\end{figure}
For evaluating the diversity of the generated graphs, we study four graph metrics: (i.) algebraic connectivity ($\lambda_2$), (AC), which reflects how well-connected a graph is, (ii.) (average) path length (PL), which is the (averaged) shortest distance between all possible pairs of vertices and indicates the efficiency of the information flow on the graph, (iii.) (global) clustering coefficient (CC), which measures how connected the vertices of a graph are to those in their neighborhood, and indicates how clustered the graph is, and (iv.) (average) betweenness centrality (BC),  which quantifies how often any vertex appears on the shortest path between other vertices, implies its function as an intermediary between different parts of the graph, and is a measure of the interconnectedness of the graph as a whole.

\begin{figure}[h!]

\includegraphics[trim = 1mm 10mm 1mm 8mm,clip, width=8.0cm, height=4.8cm]{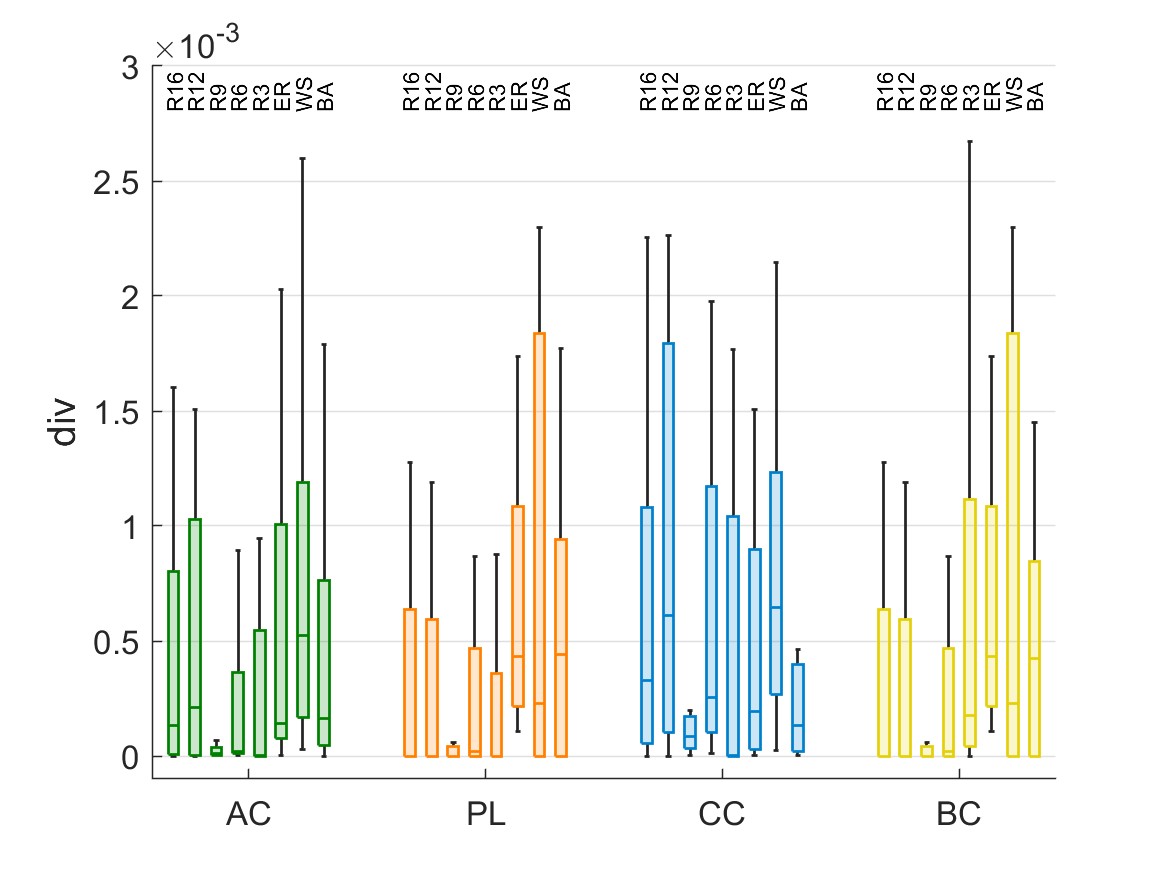}

(a)

\includegraphics[trim = 1mm 10mm 1mm 8mm,clip, width=8.0cm, height=4.8cm]{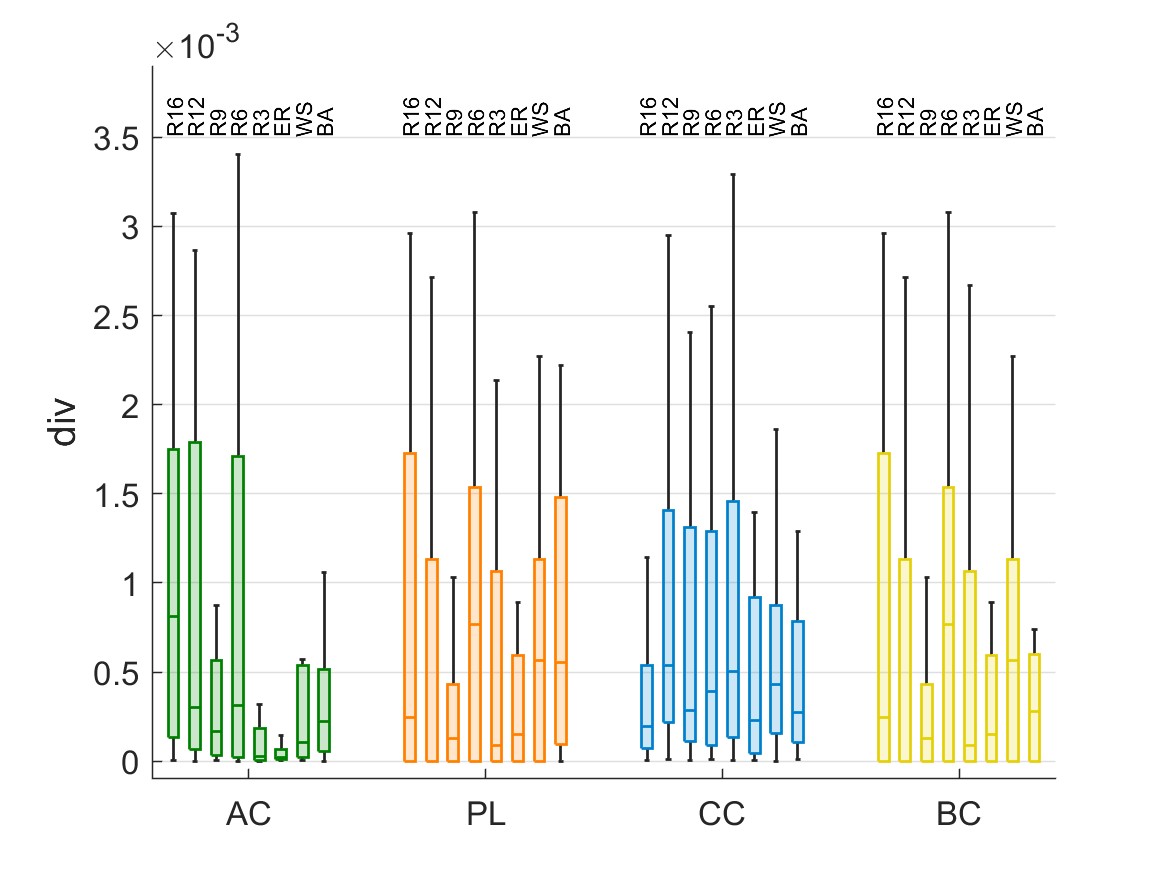}

(b) 

\includegraphics[trim = 1mm 10mm 1mm 8mm,clip, width=8.0cm, height=4.8cm]{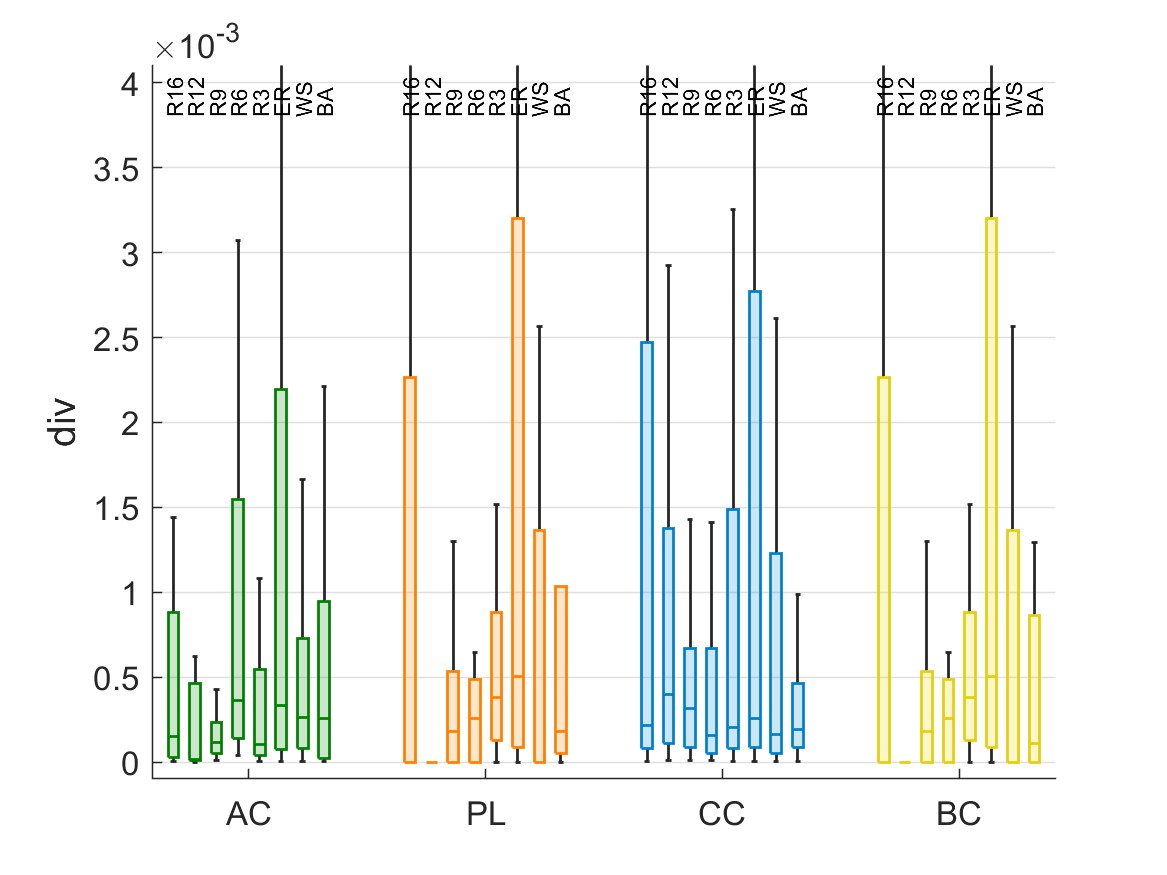}

(c)

\caption{Boxplots of the diversity contribution, see Eq. \eqref{eq:contri}. (a) Star graph $\mathcal{S}_{23}$ (b) Circulant graph $\mathcal{C}_{24}^{1,2,\ldots,8}$. (c) Circulant graph  $\mathcal{C}_{24}^{1,2,\ldots,6}$. }
\label{fig_results4}

\end{figure}

Fig.~\ref{fig_results3} shows scatter plots of
final best fitness $d$ versus the graph metrics used for quantifying diversity for the circulant target graph $\mathcal{C}_{24}^{1,2,\ldots,8}$. The experimental setup is the same as for the performance results with $n=24$ and spectral crossover 2. The black line on the level of the graph metrics is the value of the target graph.  The color code indicates different initial populations. We see that for all four graphs metrics, the values obtained for the best graphs spread about the target value. For path length (PL), clustering coefficient (CC) and betweenness centrality (BC) we find a low correlation (PL and BC positive, CC negative) to fitness. The results are similar for the other target graphs, with the exception of AC for star graph, where the target AC is the upper limit for the values obtained for best graphs.   

\subsection{Diversity of Generated Graphs}

To obtain a quantification of the diversity of the generated graphs
we apply a measure of diversity contribution~\cite{gao21} to graph metrics. Let $h(\mathcal{G}_i)$ a graph metric of graph  $\mathcal{G}_i$ and sort $h(\mathcal{G}_1)\leq h(\mathcal{G}_2)\leq \ldots h(\mathcal{G}_r) $ for $r$ samples. The diversity contribution of graph $\mathcal{G}_i$  is 
\begin{equation} \text{div}(\mathcal{G}_i)= \left(h(\mathcal{G}_i)-h(\mathcal{G}_{i-1})\right) \cdot \left(h(\mathcal{G}_{i+1})-h(\mathcal{G}_{i})\right) \label{eq:contri} \end{equation}
which depends on the next smaller and the next larger value of the graph metrics. Fig. \ref{fig_results4} shows boxplots of the four graph metrics, all considered initial populations and the target graphs. The results are again for $r=30$ runs and $1000$ generations. The graph metrics are normalized to the interval $[0,1]$ to make the diversity contributions comparable to each other. We see that in general good diversity is achieved.  It can further be observed that for all target graphs and most initial populations a satisfying amount of diversity is obtained for all considered graph metrics. However, almost all boxplots have at least whiskers that touch the zero line, which indicates that at least some evolved graphs have vanishing diversity contribution, meaning they have the same (or very similar) graph metrics as other evolved graphs. Occasionally, we obtain  results with generally small diversity contribution, for instance for star graph target,  $9$-regular input graphs and all graph metrics, or circulant graph target $\mathcal{C}_{24}^{1,2,\ldots,8}$,  ER input graphs and algebraic connectivity, or  circulant graph target $\mathcal{C}_{24}^{1,2,\ldots,6}$, $12$-regular input graphs, and path length and betweenness centrality. A comparison to performance (see Fig.~\ref{fig_results}) shows that there is no direct relationship between these results and either low or high fitness. Thus, the lack of graph diversity is most likely not linked to fitness measured by spectral distance and for remedy additional designs are needed, for instance  using the desired graph metrics as co-objective.

\section{Conclusions% and future work
}

We have presented a new evolutionary approach for evolving graphs aimed at producing networks whose Laplacian spectra match those of given target graphs. The scheme allows to generate new graphs that share high level properties with a given target.

Good results have been obtained by using mutation control through algebraic connectivity and  crossover which involves spectral clustering. Apart from matching target spectra, we have also shown that the generated graphs have diversity with respect to (non-spectral) graph metrics, such as path length, clustering coefficient and betweenness centrality. Thus, our approach is applicable to situations where graphs are needed for experimentally evaluating network algorithms and protocols and where the graphs should match the high-level connectivity information expressed by the Laplacian spectra, but be diverse in terms of non-spectral graph descriptors. 

\ignore{
In reporting first results on evolving graphs towards desired spectra, the paper inherently motivates a set of questions that point to possible directions for future research.
A first direction is enhancing control over  edge manipulations carried out by the genetic operators. The current implementation has mutation that adds or removes edges and changes degree, while crossover mostly recreates the number of edges removed by the graph partitioning involved in the crossover procedure. An additional crossover control could tune the number of edges used to join the recombined subgraphs and thus more strongly  change relative degree of offspring graphs and consequently influence the dynamics of density.  Also, 
mutation currently only adds or removes edges. Particularly if the evolving graph is close to the target spectra, controlled edge rewiring and density drift aimed at changing the local community structure of the graph might improve results. Another main result of the experiments was that some classes of initial graphs work better than others. Thus, a second direction is further study of the interplay between the design of the genetic operators on the one hand, and transitions between initial graphs and target densities on the other.   
}

\section*{Data and code availability}
 The program code and data to replicate the results 
 can be found at the data repository \url{https://github.com/HendrikRichterLeipzig/EA_graph_spectra}.

\section*{Acknowledgment} The main part of this work was done during a research visit by Hendrik Richter to the University of Adelaide, Australia. He gratefully acknowledges the hospitality of the host institution and the support of the Deutsche Forschungsgemeinschaft (DFG, German Research Foundation) through grant No. 558767281.

\balance


\begin{thebibliography}{1}



\bibitem{alb02}  Réka Albert  and Albert-László Barabási. 2002.  Statistical mechanics of complex networks, Rev. Mod. Phys. 74, 47. 

\bibitem{ash14} Daniel Ashlock, Justin Schonfeld, Lee-Ann Barlow, and Colin  Lee. 2014. Test problems and representations for graph evolution. In 2014 IEEE Symposium on Foundations of Computational Intelligence (FOCI), 38-45.

\bibitem{atkin18}
Timothy Atkinson,  Detlef Plump and Susan Stepney. 2018. Evolving graphs by graph programming. Proc. European Conference on Genetic Programming. 35-51. 

\bibitem{bach13} Benjamin Bach, Andre Spritzer, Evelyne Lutton, and Jean-Daniel Fekete. 2013. Interactive random graph generation with evolutionary algorithms. In Graph Drawing: 20th International Symposium, GD 2012, Redmond, WA, USA, September 19-21, 2012, Revised Selected Papers 20, 541-552. 

 \bibitem{ban08} Anirban Banerjee and Jürgen Jost. 2008. On the spectrum of the normalized graph Laplacian.
Linear Algebra Appl.  428,  3015-3022.

\bibitem{ban09}  Anirban Banerjee and Jürgen Jost. 2009.
Graph spectra as a systematic tool in computational biology.
Discrete Applied Mathematics 157 (10), 2425-2431.

\bibitem{boni20}
Angela Bonifati, Irena Holubová, Arnau Prat-Pérez, and Sherif  Sakr. 2020. Graph generators: State of the art and open challenges. ACM computing surveys (CSUR), 53(2), 1-30.

\bibitem{cir22} Petar Ćirković, Predrag Đorđević, Miloš Milićević, and Tatjana Davidović. 2022. Metaheuristic approach to spectral reconstruction of graphs. International Conference on Mathematical Optimization Theory and Operations Research. 79-93.


\bibitem{deabr07} Nair Maria Maia De Abreu. 2007. Old and new results on algebraic connectivity of graphs. Linear Algebra  Appl. 423, 53-73.

\bibitem{de18} Kevin Deweese and John R. Gilbert. 2018. Evolving difficult graphs for Laplacian solvers. In 2018 Proceedings of the Seventh SIAM Workshop on Combinatorial Scientific Computing, 35-44.

\bibitem{er59}  Paul Erdös and Alfréd  Rényi. 1959. On random graphs I, Publicationes Mathematicae, 6(26), 290–297.

\bibitem{fa21} Faezeh Faez, Yassaman Ommi,  Mahdieh  Soleymani Baghshah, and Hamid R. Rabiee. 2021. Deep graph
generators: A survey. IEEE Access 9, 106675–106702.

\bibitem{gao21}
Wanru Gao, Samadhi Nallaperuma, and Frank Neumann. 2021. Feature-based diversity optimization for problem instance classification. Evolutionary Computation  29(1), 107–128.

\bibitem{glo00} Al Globus, Sean Atsatt, John Lawton, and Todd  Wipke. 2000. JavaGenes: Evolving graphs with crossover. Technical Report. NASA Advanced Supercomputing (NAS) Division. https://www.nas.nasa.gov/assets/pdf/techreports/2000/nas-00-018.pdf

 \bibitem{gu16} Jiao Gu, Jürgen Jost, Shiping Liu, and Peter F. Stadler. 2016. Spectral classes of regular, random, and empirical graphs.
Linear Algebra Appl. 489, 30--49.

\bibitem{guo22} Xiaojie Guo and Liang  Zhao. 2022. A systematic survey on deep generative models for graph generation. IEEE Transactions on Pattern Analysis and Machine Intelligence, 45(5), 5370-5390.

\bibitem{hig07} Desmond J. Higham,  Gabriela Kalna, and Milla Kibble. 2007. Spectral clustering and its use in bioinformatics. Journal of Computational and Applied Mathematics 204.1, 25-37.

\bibitem{ibs02} Mads Ipsen and Alexander S. Mikhailov. 2002. Evolutionary reconstruction of networks. Phys. Rev. E 66.4, 046109.

\bibitem{jia14} Hongjie Jia, Shifei  Ding, Xinzheng Xu, and Ru Nie. 2014. The latest research progress on spectral clustering. Neural Computing and Applications, 24(7), 1477-1486.

\bibitem{jim21} Aarón Jiménez-Aparicio,  Efrén Mezura-Montes, and Héctor-Gabriel Acosta-Mesa. 2021. Evolutionary algorithms for searching almost-equienergetic graphs. In 2021 IEEE Congress on Evolutionary Computation (CEC), 1093-1098.

\bibitem{kim11} Jin Kim, Inwook Hwang, Yong-Hyuk Kim, and Byung-Ro  Moon. 2011. Genetic approaches for graph partitioning: a survey. In Proceedings of the Genetic and Evolutionary Computation Conference,  473-480.

\bibitem{leith24} Sydney Leither,  Vincent Ragusa, and Emily Dolson. 2024. Evolving weighted and directed graphs with constrained properties. In Proceedings of the Genetic and Evolutionary Computation Conference Companion,  443-446.

\bibitem{liu19} Yang Liu,  Xi Wang, and Jürgen Kurths. 2019. Framework of evolutionary algorithm for investigation of influential nodes in complex networks. IEEE Transactions on Evolutionary Computation, 23(6), 1049-1063.

\bibitem{loz21} Manuel Lozano  and Francisco J. Rodriguez. 2021. Network reconstruction from betweenness centrality by artificial bee colony. Swarm and Evolutionary Computation, 62, 100851.

\bibitem{luo23} Tianze Luo,  Zhanfeng Mo, and Sinno Jialin Pan. 2023. Fast graph generation via spectral diffusion. IEEE Transactions on Pattern Analysis and Machine Intelligence 46.5, 3496-3508.

 
\bibitem{martin06} Jacob G. Martin. 2006. Spectral techniques for graph bisection in genetic algorithms. In Proceedings of the  Genetic and Evolutionary Computation Conference, 1249-1256.

\bibitem{med23} Eric Medvet,  Simone Pozzi, and Luca Manzoni. 2023. A general purpose representation and adaptive EA for evolving graphs. In Proceedings of the Genetic and Evolutionary Computation Conference, 1156-1164.

\bibitem{mer99} Markus Meringer. 1999. Fast generation of regular graphs and construction of cages. J. Graph Theory 30, 137-146.

\bibitem{mon02} Oleg  Monakhov and Emilia Monakhova. 2002. Using evolutionary algorithm for generation of dense families of circulant networks. In Proceedings of the 2002 Congress on Evolutionary Computation. CEC'02, 1854-1859.

\bibitem{mur24} Bogdan-Eduard-Mădălin Mursa and Anca Andreica. 2024. Generating random complex networks with network motifs using evolutionary algorithm-based null model. Swarm and Evolutionary Computation
86, 101526.

\bibitem{papa21}
Evgenia Papavasileiou,  Jan Cornelis, and Bart Jansen. 2021. A systematic literature review of the successors of "neuroevolution of augmenting topologies. Evolutionary Computation 29(1), 1-73.

\bibitem{rich21} Hendrik Richter. 2021. Spectral analysis of transient amplifiers for death-birth updating constructed from regular graphs. J. Math. Biol. 82, 61.


\bibitem{rich23} Hendrik Richter. 2023. Spectral dynamics of guided edge removals and identifying transient amplifiers for death-Birth updating. J. Math. Biol. 87, 3.

\bibitem{shi00}  Jianbo Shi and Jitendra Malik. 2000. Normalized cuts and image segmentation. IEEE Transactions on Pattern
Analysis and Machine Intelligence, 22(8), 888-905.

\bibitem{shi19}
Alana Shine and David Kempe. 2019. Generative graph models based on Laplacian spectra? In: Liu, L., White, R. (eds) WWW'19: The World Wide Web Conference,     ACM, New York,  1691-1701.

\bibitem{sotto20} Léo Françoso D. P. Sotto, Paul Kaufmann, Timothy Atkinson, Roman Kalkreuth, and  Márcio Porto Basgalupp. 2020. A study on graph representations for genetic programming. In Proceedings of the 2020 Genetic and Evolutionary Computation Conference, 931-939.

\bibitem{th22} Henri Th\"olke and Jens  Kosiol. 2022. A multiplicity-preserving crossover operator on graphs. Proc. 25th International Conference on Model Driven Engineering Languages and Systems, 588-597. 


\bibitem{ver17} Merijn Verstraaten, Ana Lucia Varbanescu, and Cees de Laat.  2017. Synthetic graph generation for systematic exploration of graph structural properties. In: Desprez, F., et al. Euro-Par 2016: Parallel Processing Workshops. Euro-Par 2016. Springer. 

\bibitem{watts98}  Duncan J. Watts and Steven H. Strogatz. 1998. Collective dynamics of ‘small-world’
networks. Nature 393, 440–442.



\bibitem{wills20}
Peter Wills and François G. Meyer.  2020. Metrics for graph comparison: A practitioner’s guide. PLoS ONE 15(2), e0228728.
































































\end{thebibliography}
\end{document}